\documentclass{article} % For LaTeX2e

\usepackage{times}

\usepackage[a4paper, margin=1in]{geometry}

\usepackage[utf8]{inputenc} % allow utf-8 input
\usepackage[T1]{fontenc}    % use 8-bit T1 fonts

\usepackage{url}            % simple URL typesetting
\usepackage{booktabs}       % professional-quality tables
\usepackage{amsfonts}       % blackboard math symbols
\usepackage{nicefrac}       % compact symbols for 1/2, etc.
\usepackage{microtype}      % microtypography
\usepackage{xcolor}         % colors
\usepackage[pdftex]{graphicx}
\usepackage{algorithm}
\usepackage[noend]{algpseudocode}
\usepackage{graphicx}
\usepackage{doi}
\usepackage{amsmath}
\usepackage{amssymb}
\usepackage{mathtools}
\usepackage{amsthm}

\usepackage{enumitem}
\usepackage{natbib}
\usepackage{lipsum}

\def\gC{{\mathcal{C}}}

\def\gM{{\mathcal{M}}}

\def\rmA{{\mathbf{A}}}

\def\rmG{{\mathbf{G}}}

\def\rvb{{\mathbf{b}}}
\def\rvc{{\mathbf{c}}}

\newtheorem{defn}{Definition}

\newtheorem{example}{Example}

\usepackage[capitalize]{cleveref}
\crefname{thm}{Theorem}{Theorems}
\crefname{lem}{Lemma}{Lemmas}
\crefname{cor}{Corollary}{Corollaries}
\crefname{prop}{Proposition}{Propositions}
\crefname{asmp}{Assumption}{Assumptions}
\crefname{defn}{Definition}{Definitions}
\crefname{oracle}{Oracle}{Oracles}
\crefname{fact}{Fact}{Facts}
\crefname{conj}{Conjecture}{Conjectures}
\crefname{rem}{Remark}{Remarks}
\crefname{example}{Example}{Examples}
\crefname{condition}{Condition}{Conditions}
\crefname{exercise}{Exercise}{Exercises}
\crefname{algorithm}{Algorithm}{Algorithms}
\crefname{table}{Table}{Tables}
\crefname{figure}{Figure}{Figures}
\crefname{section}{Section}{Sections}
\crefname{subsection}{Section}{Sections}
\crefname{appendix}{Appendix}{Appendices}
\crefname{mess}{Message}{Messages}
\crefname{claim}{Claim}{Claims}
\crefname{question}{Question}{Questions}
\crefname{case}{Case}{Cases}

\usepackage{listings}
\usepackage{tcolorbox}
\usepackage{multirow}
\usepackage{hyperref}
\hypersetup{
    colorlinks=true,
    citecolor=blue,
    linkcolor=blue,
}

\theoremstyle{plain}
\newtheorem{theorem}{Theorem}[section]

\newtheorem{lemma}[theorem]{Lemma}

\theoremstyle{definition}
\newtheorem{definition}{Definition}[section]

\theoremstyle{remark}

\newcommand{\sectionspace}{\vspace{-0.2cm}}

\title{ORGEval: Graph-Theoretic Evaluation of LLMs in
Optimization Modeling}

\author{
\textbf{Zhuohan Wang}$^{*1}$ \quad \textbf{Ziwei Zhu}$^{*1}$ \quad \textbf{Ziniu Li}$^{1}$ \quad \textbf{Congliang Chen}$^{23}$ \\
\textbf{Yizhou Han}$^{1}$ \quad \textbf{Yufeng Lin}$^{1}$ \quad \textbf{Zhihang Lin}$^{1}$ \quad \textbf{Angyang Gu}$^{1}$ \\
\textbf{Xinglin Hu}$^{1}$ \quad \textbf{Ruoyu Sun}$^{14}$ \quad \textbf{Tian Ding}$^{\dagger24}$ \\[6pt]
$^{1}$The Chinese University of Hong Kong, Shenzhen \\
$^{2}$Shenzhen Research Institute of Big Data \\
$^{3}$Shenzhen Loop Area Institute \\
$^{4}$Shenzhen International Center for Industrial and Applied Mathematics \\[6pt]
\textit{*: Equal contribution; $\dagger$: Correspondence author.}
}

\begin{document}

\maketitle

\begin{abstract}
%In Operations Research (OR), 
Formulating optimization problems for industrial applications demands significant manual effort and domain expertise. While Large Language Models (LLMs) show promise in automating this process, evaluating their performance remains difficult due to the absence of robust metrics. Existing solver-based approaches often face inconsistency, infeasibility issues, and high computational costs. To address these issues, we propose ORGEval, a graph-theoretic evaluation framework for assessing LLMs’ capabilities in formulating linear and mixed-integer linear programs. ORGEval represents optimization models as graphs, reducing equivalence detection to graph isomorphism testing. We identify and prove a sufficient condition, when the tested graphs are symmetric decomposable (SD), under which the Weisfeiler–Lehman (WL) test is guaranteed to correctly detect isomorphism. Building on this, ORGEval integrates a tailored variant of the WL-test with an SD detection algorithm to evaluate model equivalence. By focusing on structural equivalence rather than instance-level configurations, ORGEval is robust to numerical variations.  Experimental results show that our method can successfully detect model equivalence and produce 100\% consistent results across random parameter configurations, while significantly outperforming solver-based methods in runtime, especially on difficult problems. Leveraging ORGEval, we construct the Bench4Opt dataset and benchmark state-of-the-art LLMs on optimization modeling. Our results reveal that although optimization modeling remains challenging for all LLMs, DeepSeek-V3 and Claude-Opus-4 achieve the highest accuracies under direct prompting, outperforming even leading reasoning models.

\end{abstract}

%%%%%%%%%%%%%%%%%%%%%%%%%%%%%%%%%%%%%%%%%%%%%%%%%%%%%%%%%%%%%%%%%%%%%%%%%%%%%%%%%%%%%%%%%%%
%%intro%%%%%%%%%%%%%%%%%%%%%%%%%%%%%%%%%%
%%%%%%%%%%%%%%%%%%%%%%%%%%%%%%%%%%%%%%%%%%%%%%%%%%%%%%%%%%%%%%%%%%%%%%%%%%%%%%%%%%%%%%%%%%%

\vspace{-2mm}
\section{Introduction}
\sectionspace 

Operations Research (OR) leverages mathematical modeling and optimization techniques to support decision making in complex systems \citep{hillier2015introduction}. It plays a vital role across industries such as logistics, manufacturing, finance, and healthcare \citep{winston2004operations}, where real-world scenarios are formalized into mathematical optimization models. However, this formulation process is highly challenging, demanding interdisciplinary expertise. Recently, there has been growing interest in using Large Language Models (LLMs) to automate the formulation of optimization models from user inputs, write proper programs, and solve complex problems with minimal human intervention \citep{jiang2024llmopt,huang2025orlm,lu2025optmath}.

Evaluating modeling correctness is challenging because fundamentally equivalent optimization models can be represented by different variable names and structures. To handle this non-uniqueness, prior work has primarily adopted solver-based evaluation, which solves the generated model to obtain its objective value and then compares it against the ground-truth optimal objective \citep{tang2024orlm}. The assumption is that correct models should produce matching objective values. However, we noticed that solver-based evaluation suffers from a few major limitations: 1) models may coincidentally have the same optimal value under one parameter configuration but produce distinct values under another, 2) the solver fails to evaluate model equivalence when input parameters result in an infeasible problem, and 3) solvers may encounter high computation costs for a single round evaluation. Moreover, current evaluation settings focus on problems that embed both the description and the numerical data within a single prompt \citep{ramamonjison2022augmenting,xiao2023chain}, which often constrains the problem size to relatively small instances. This approach can not reflect many real-world scenarios, where data is typically large-scale and is separated from the modeling process \citep{apio9th}.  
%that may significantly hinder the large-scale application of LLMs for OR:

\begin{figure}[t]
\begin{center}
%\framebox[4.0in]{$\;$}
%\fbox{\rule[-.5cm]{0cm}{4cm} \rule[-.5cm]{4cm}{0cm}}
\includegraphics[width = \textwidth]{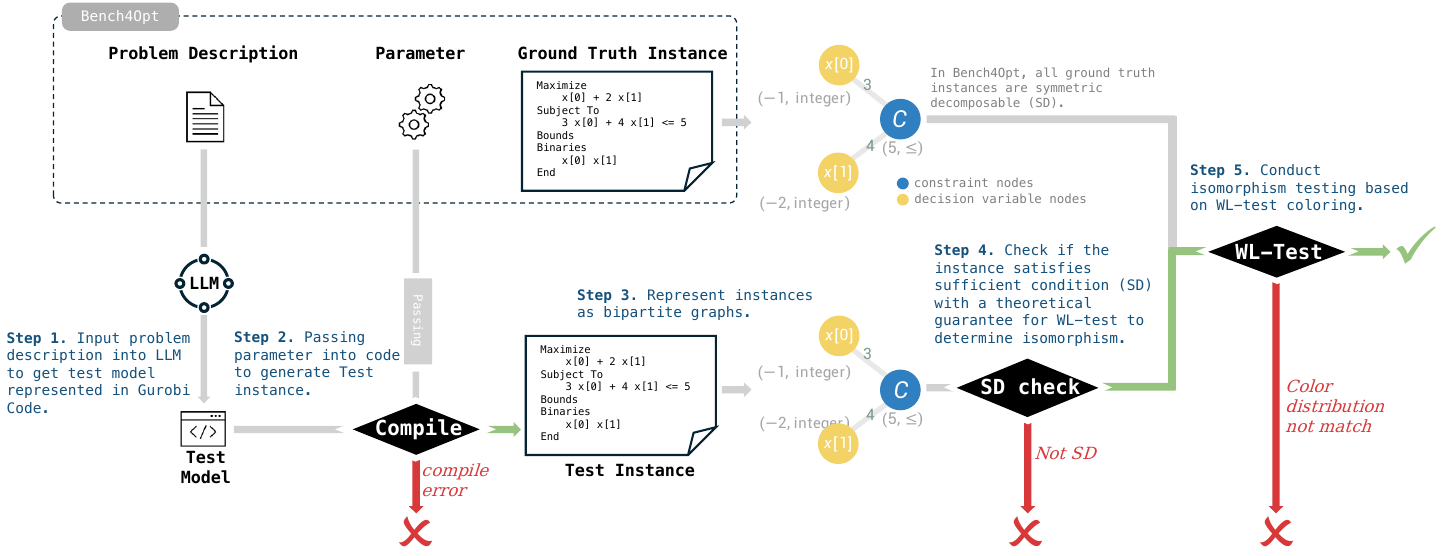}
\end{center}
\caption{Evaluation Pipeline: Each example in our dataset includes a problem description, a parameter file, and a model instance with parameters applied. To assess an AI system's modeling capability, we evaluate the equivalence between the AI-generated instance and the ground truth instance in our dataset, using a common set of parameters. These instance pairs can be represented by two bipartite graphs, on which we applied an isomorphism testing algorithm, and meanwhile, checked the sufficiency of the algorithm.}
\label{figure:intro}  
\end{figure}

In this paper, we address these limitations by introducing ORGEval, a novel evaluation method for assessing modeling equivalence and correctness for linear and mixed-integer linear programs—the dominant classes in practical OR applications. Specifically, we formulate optimization models as graphs, then reduce the equivalence detection task to a graph isomorphism problem. We identify sufficient conditions, symmetric decomposable graphs, under which the Weisfeiler-Lehman (WL) test is guaranteed to detect isomorphism. Additionally, we develop a selection algorithm that identifies problems satisfying these conditions
and combined this selection algorithm to develop an enhanced WL test that checks isomorphism.

% We identify sufficient conditions—WL-determinable and symmetric decomposable graphs—that guarantee the Weisfeiler–Lehman (WL) test can correctly detect isomorphism. We further develop an algorithm to select problems satisfying these conditions and propose an enhanced WL test that incorporates additional steps to check whether these sufficient conditions for isomorphism detection are met.

Our main contributions can be summarized as follows:
\vspace{-2mm}
\begin{enumerate}
    \item We formalize model equivalence under the model–data separation setting (Section \ref{section:problem_setting}).
    \item We reduce model equivalence detection into graph isomorphism testing (Section \ref{subsection:evaluation_principle}).
    \item We identify and prove a sufficient condition that enhances existing graph isomorphism testing algorithms (Section \ref{subsection:graph_iso} and Section \ref{subsection:sufficient_condition}).
    \item We propose ORGEval, a graph-theoretic evaluation method for assessing optimization models (Section \ref{subsection:orgeval}).
    \item We introduce Bench4Opt, the first model–data separated dataset for optimization modeling. Using Bench4Opt, we empirically demonstrate the efficiency and consistency of ORGEval, and further benchmark the performance of leading LLMs (Section \ref{section:exp}).
\end{enumerate}

\vspace{-2mm}
\sectionspace 
\section{Problem Formulation}
\label{section:problem_setting}
\sectionspace 

In this section, we establish the background for the OR modeling task and define the capabilities we aim to measure and evaluate. First, we note that real-world OR modeling typically unfolds through three distinct stages:
\begin{itemize}[topsep=1pt,parsep=1pt,partopsep=1pt, leftmargin=*]
    \item  \textbf{Pre-modeling stage:} Stakeholders articulate problems in natural language (e.g., "I want to maximize revenue through car production") and collect all relevant numerical parameters.
    \item \textbf{Modeling stage:} Analysts transform these natural language descriptions into mathematical models by defining decision variables, formulating objective functions, and establishing constraints. This results in mathematical formulations or solver-ready code with separately stored parameters.
    \item \textbf{Post-modeling stage:} The model is instantiated with problem-specific data to generate a fully specified problem ready for computational solving.
\end{itemize}
We explicitly distinguish between the modeling stage and the post-modeling stage. In many real-world applications, models are designed for reusability: the structural components (decision variables, objectives, and constraints) are specified once, while the problem-specific parameters are supplied separately for each instance. This separation is standard practice in modern modeling languages such as AMPL \citep{ampl-data} or Pyomo \citep{hart2011pyomo}, where the model (.mod) and the data (.dat) are maintained as distinct artifacts. We thus view “instantiation with problem-specific data” as a necessary step that bridges abstract modeling and computational solving.
\begin{figure}[t]
\begin{center}
%\framebox[4.0in]{$\;$}
%\fbox{\rule[-.5cm]{0cm}{4cm} \rule[-.5cm]{4cm}{0cm}}
\includegraphics[width = 0.8\textwidth]{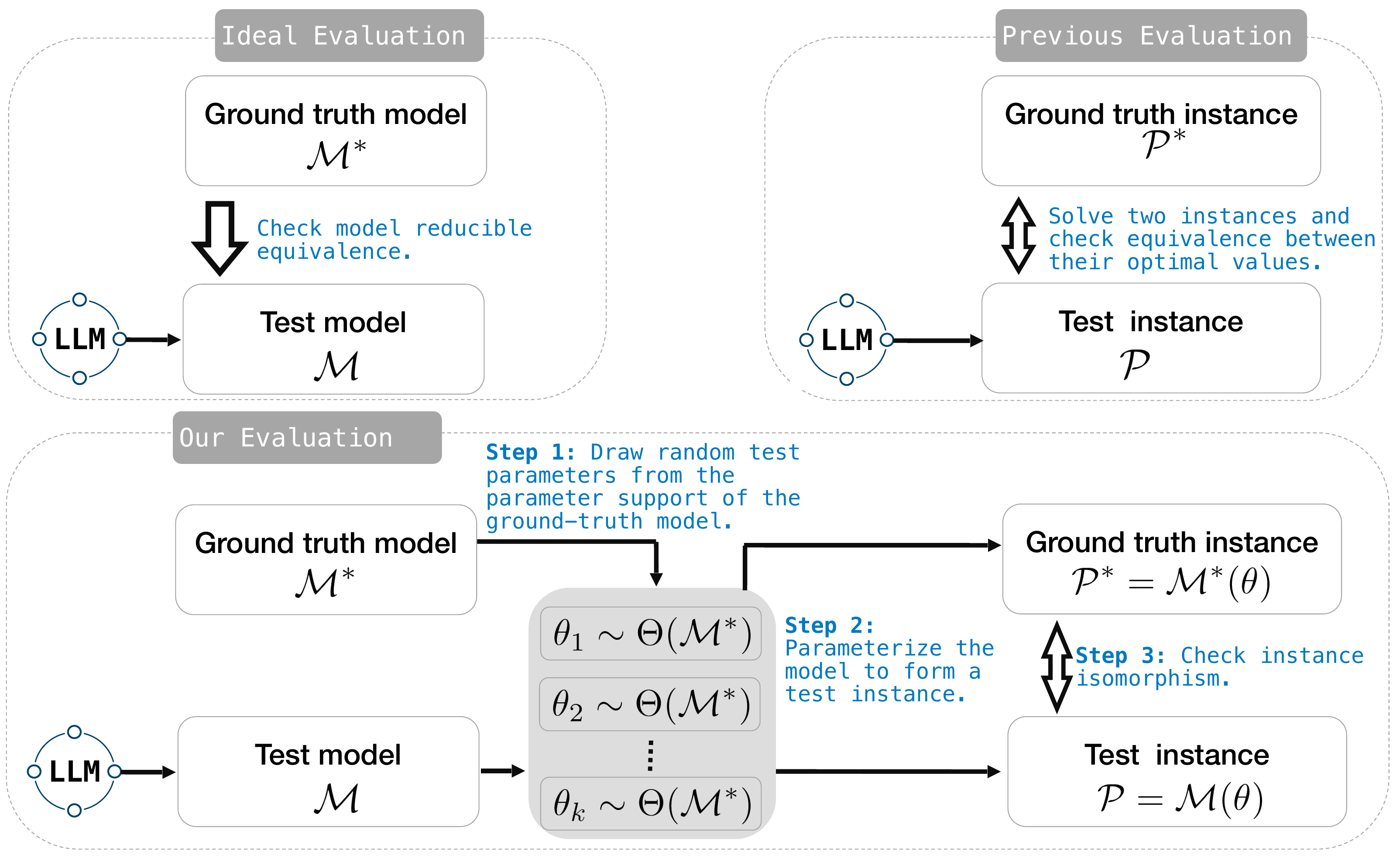}
\end{center}
\caption{Evaluation Framework: The ultimate goal of modeling equivalence is to directly assess whether one model can be equivalently transformed to a standard model (top left). Existing work tests the equivalence between numerical instances by comparing their optimal objective (top right). Our evaluation method approximates the ultimate goal of directly evaluating modeling equivalence by randomly sampling instances and testing instance isomorphism (bottom).}
\label{figure:evaluation_framework}  
\end{figure}

Our primary goal is to leverage LLMs to automate the first two stages, enabling users to input problem descriptions and receive both mathematical formalizations and solver-ready code as outputs. We formally reframed the evaluation of modeling equivalence in the following subsections and illustrated ideal evaluation, previous evaluation, and our new evaluation framework in Figure \ref{figure:evaluation_framework}.

\sectionspace 
\subsection{Definition of Modeling "Accuracy"}
\sectionspace 

% % distinguishing between different forms of structural consistency between models.

Our work proposes a formal evaluation framework for model-level "accuracy", which requires the definition of model equivalence. 
To support the definition of this kind of equivalence, 
we introduce the notions of \emph{modeling problem instance} and \emph{modeling parameter support}. 

\subsubsection{Definitions for Modeling Problems}
\begin{defn}[Modeling Problem Instance]
A MILP/LP problem instance, denoted by $\mathcal{P}$, has the following standard formulation \citep{luenberger1984linear}: 
\begin{align}  \label{equation:eq1}
\quad \min_{\mathbf{x}\in \mathbb{R}^p\times \mathbb{Z}^{n-p}} \quad & \mathbf{c}^\intercal\mathbf{x}, \\
\text{such that} \quad  &\mathbf{A} \mathbf{x} \circ \mathbf{b},  \nonumber
\end{align}
where $\mathbf{A}\in \mathbb{R}^{m\times n}$, $\mathbf{c}\in \mathbb{R}^n$, $\mathbf{b}\in \mathbb{R}^m$, and $\circ\in \{=,<,>,\leq,\geq\}^m$ for $i = 1, \cdots, n$. Note that in this formulation, there are $p$ real optimization variables and $n - p$ integer optimization variables, and $m$ constraints.
\end{defn}
To facilitate later usage, we use a vector $\mathbf{\tau}$ to represent the decision variable type, where $\tau_i =1$ indicates $x_i$ is an integer and $\tau_i =0$ indicates a continuous variable.

\begin{defn}[MILP/LP Model]
A MILP/LP model is a mapping $\mathcal{M}: \Theta \rightarrow \mathcal{P}$, where $\Theta$ denotes a space of model parameters. Each $\theta \in \Theta$ is a tuple $\theta = (A, c, \tau, b, \circ)$,
where $A, c, b, \circ$ are the same as in Definition~1, and $\tau \in \{0,1\}^n$ is a binary vector indicating whether each variable is continuous ($0$) or integer ($1$). 
\end{defn}

We refer to $\Theta$ as the \textbf{modeling parameter support} of $\mathcal{M}$. Given any $\theta \in \Theta$, the mapping $\mathcal{M}(\theta)$ returns a concrete MILP or LP instance $\mathcal{P}$ of the form~\cref{equation:eq1}. An example of a blending optimization model $\mathcal{M}_{blend}$ and its parameter support $\Theta(\mathcal{M}_{blend})$ can be found in \cref{example:model_parameter_set} in the appendix. Figure \cref{figure:model_instance_generation} illustrates how to generate model instances for $\mathcal{M}_{blend}$.

\vspace{-2mm}
\subsubsection{Definitions for Modeling Equivalence}
In practical optimization workflows, "modeling equivalence" should reflect whether one predicted model can be systematically transformed to a target model. 
\begin{defn}[Model-lossless-reduction]
\label{definition: model_reduction}
For two models $\gM_1$ and $\gM_2$, they are said to be model-lossless-reducible if the following conditions hold:
\begin{enumerate}
\vspace{-2mm}
\item Shared parameter support: $\mathcal{M}_1$ and $\mathcal{M}_2$ share the same parameter support $\Theta$;
\vspace{-2mm}
\item Existence of solution-preserving transformation: There exists a mapping $F$ over decision variables such that, for any parameter $\theta \in \Theta$:
\begin{itemize}
\vspace{-2mm}
    \item If $\mathcal{M}_1(\theta)$ is feasible and bounded, then $\mathcal{M}_2(\theta)$ is also feasible and bounded, and for any optimal solution $x^*$ of $\mathcal{M}_1(\theta)$, $F(x^*)$ is an optimal solution of $\mathcal{M}_2(\theta)$;
    \item If $\gM_1(\theta)$ is infeasible or unbounded, then $\gM_2(\theta)$ is also infeasible or unbounded.
    
\end{itemize}

\end{enumerate}

\end{defn}
This model-lossless-reduction captures the essence of equivalence between models: if one model can fully simulate the feasible behaviors and optimal outcomes of another, we treat them as equivalent for all practical purposes. 

However, such a form of equivalence is difficult to check and has not been reliably captured by existing works. 
First, it is hard to verify the equivalence between parameter support. Second, verifying solution mappings $F$ between models can be computationally expensive and sometimes impractical, as it may require comparing optimality across a large space of instances rather than a single solution.

As verifying model-lossless-reducibility is often intractable in practice, current approaches typically rely on a much weaker proxy, comparing solver outputs on specific model instances. In the following section, we will formally introduce solver-based modeling accuracy evaluation and discuss its limitations.

\sectionspace 
\subsection{Solver-based Modeling "Accuracy"}
\sectionspace 
%When benchmarking LLMs' capabilities for OR modeling tasks, we need to measure the "accuracy" of LLM-generated models. This presents a scientific and technical challenge: how should we define "accuracy" for LLM-generated modeling? This usually lies in the LLM-generated code for the mathematical model. It is clear that direct comparison of LLM-generated code with ground truth code is ineffective because variable names and data structures can vary significantly.
Previous work \citep{tang2024orlm} explored evaluating modeling "accuracy" through execution accuracy, which we formalize below:
\begin{defn}[Execution Accuracy]
For a data configuration $\theta$, a mathematical model $\gM$ with a program code $\gC$ is said to be correct if, upon execution of $\gC$, we obtain $z(\gC, \theta) = z^{\star}$, where $z^{\star}$ is the optimal value of the mathematical model if it exists.
\end{defn}
We point out several limitations of this definition. 
% when we formally write the definition of execution accuracy.
\begin{itemize}[topsep=1pt,parsep=1pt,partopsep=1pt, leftmargin=*]
\item \textbf{(Limitation 1) No rigorous correctness guarantee when the solver returns values:} Cases exist where final answers appear correct despite fundamentally flawed underlying optimization models; see example \ref{example:limitation1_example1} and example \ref{example:limitation1_example2} in Appendix \ref{appendix: examples}.

\item \textbf{(Limitation 2)  Uninformative in cases of infeasible problems} Mathematical models could be \emph{infeasible} under certain data configurations, in which cases the solver would return no feasible solution. Solver-based evaluation becomes uninformative in this scenario; see \ref{example:limitation2_example1} in Appendix \ref{appendix: examples}. 

\item \textbf{(Limitation 3) Prohibitively high computational costs:} Solvers may require hundreds and thousands of CPUs and run for several hours and days, especially for large-scale problems such as Mixed Integer Linear Programming (MILP) tasks. This makes the evaluation time-consuming and computationally expensive. This would further make it impractical to apply advanced post-training techniques like those proposed by \citep{o1openai, guo2025deepseek} that could otherwise enhance LLM performance.

\end{itemize}
Fundamentally, the limitations illustrated above reveal a critical limitation in the execution accuracy definition: it is restricted to a \emph{single} data configuration and relies solely on optimal value comparison. This approach is clearly inadequate for reliable model evaluation. A mathematical model should be correct across \emph{all} possible data configurations, not just one specific test case. Only when a model demonstrates consistent performance across diverse scenarios can we trust the decision information it provides for practical applications. Therefore, an ideal evaluation metric for model equivalence is expected to be reliable, informative, and consistent over various data configurations.

% \begin{table}[]
%     \centering
%     \begin{tabular}{c|c}
%     \toprule 
%        Notation  &  Meaning \\  \hline 
%        $\gP$          & an LP/MILP problem instance \\
%        $\theta$       & data parameters values (e.g., cost vector $c$, constraint matrix $A$, vector $b$) \\
%        $\mathbf{G}$   & bipartite graph representation of LP/MILP \\
%        $\gM$          & an   \\ 
%        \bottomrule
%     \end{tabular}
%     \caption{Caption}
%     \label{tab:my_label}
% \end{table}

% We focus on the modeling stage and target to assess equivalence between models, which serves as a critical bridge between abstract reasoning and computational implementation. Our goal is to evaluate the equivalence of two models, in the sense that they represent the same underlying logic or problem structure, instead of the functional equivalence of two specific model instances. This is important because once a robust model is developed in the modeling stage, it can be reused and adapted as real-world data changes in the future, significantly improving the model's longevity and scalability. 

\subsection{Model Isomorphism}
As an alternative, we further define model isomorphism to capture structural equivalence between models, rather than relying on numerical optimal values.
\begin{defn}[Model Isomorphism] 
We say two optimization models $\mathcal{M}_1$ and $\mathcal{M}_2$ are model isomorphic if the following conditions hold: 
\begin{itemize}
\vspace{-2mm}
    \item Shared parameter support: $\mathcal{M}_1$ and $\mathcal{M}_2$ share the same parameter support $\Theta$;
    \vspace{-2mm}
    \item Existence of permutation-invariant transformation: There exists a permutation mapping $F_1$ over decision variables and $F_2$ over constraints such that, for any $\theta \in \Theta$, the transformed instance $F_1\circ F_2(\mathcal{M}_1(\theta))$ is exactly $\mathcal{M}_2(\theta)$.
\end{itemize}
\end{defn}

\vspace{-2mm}
In this work, we refer to this as isomorphism equivalence, or simply equivalence. Note that model isomorphism is sufficient for mutual reducibility: if $\mathcal{M}_1$ and $\mathcal{M}_2$ are isomorphic, then $\mathcal{M}_1$ and $\mathcal{M}_2$ are mutually model-lossless-reducible.

%Under the definition of model isomorphism, we can naturally define instance-level isomorphism:

% instance-level isomorphism 可以在下一个section定义，二分图转换也放在下一个section
% section2: 1. 正确建模 正确应该是什么2. 现在的工作做不到，3. 我们提出一个sufficient condition
\vspace{-2mm}
\paragraph{Optimization Model Equivalence}
Previous work typically assesses instance-level equivalence by comparing the optimal values of $\mathcal{P}_1 = \mathcal{M}_1(\theta)$ and $\mathcal{P}_2 = \mathcal{M}_2(\theta)$ for specific $\theta$, where both the model and data are coupled in the input prompt. Different from previous work, we aim to detect model-level equivalence and consider a model-data separated framework for evaluating autonomous modeling systems. To make model-level equivalence evaluation tractable, we reduce model "equivalence" to model isomorphism, a stricter form of equivalence that focuses on the inherent structure of models rather than their optimal solutions. Importantly, we observe that if two instances are isomorphic, their isomorphism remains unchanged under different data configurations. This property allows us to efficiently evaluate equivalence at the model level without repeatedly solving individual model instances.

Specifically, we evaluate equivalence between $\mathcal{M}_1$ and $\mathcal{M}_2$ by randomly sampling parameters $\mathbf{\theta}$ from $\Theta(\mathcal{M})$ and testing the isomporphism of modeling instances $\mathcal{M}_1(\mathbf{\theta})$ and $\mathcal{M}_2(\mathbf{\theta})$. This serves as an empirical approximation of model-level correctness.

\sectionspace 
\section{Methodology}
\label{section:evaluation}
\vspace{-2mm}
In this section, we introduce our evaluation method ORGEval, and the theoretical guarantee of ORGEval. ORGEval is developed to evaluate modeling equivalence by detecting whether the inherent structure of two random instances from two models is equivalent, thus making the evaluation stable when altering instance parameters.

\subsection{Evaluation Principle}
\label{subsection:evaluation_principle}

%We propose a new evaluation paradigm, which identifies the correctness of optimization modeling by detecting whether the inherent structure of a \emph{test instance} is isomorphic to that of the \emph{standard instance}, thus making the evaluation stable when altering instance parameters.

%We first establish the correctness of MILP and LP model formulation based on the following three principles: (1) Variables should reflect real-world entities. (2) Objectives should clearly align with their descriptions, and (3) Constraints should represent real-world limitations without redundancy. %A model following the above principles should be regarded as correct. 

We first introduce our notion of \textbf{equivalence} between two model instances. Specifically, our notion allows model instances to change the notations of variables and rearrange their variables/constraints in different orders without losing essential information. The formal definition is as follows:
\begin{defn}[Instance-Level Isomorphism] \label{definition:instance_equivalent}
We say model instances $\mathcal{P}_2$ and $\mathcal{P}_1$ are isomorphic, if $\exists$ permutation matrices $P_1, P_2$ which shuffle the index of a vector or column index of a matrix such that $\mathcal{P}_2$ can be written in the following form:
\vspace{-2mm}
\begin{align*}
\min_{\mathbf{x}\in \mathbb{R}^{n-p}\times \mathbb{Z}^p}& \hat{\mathbf{c}}^T\mathbf{x}, \\
\text{s.t. }& \hat{\mathbf{A}} \mathbf{x} \, \hat{\circ} \, \hat{\mathbf{b}}, 
\end{align*}

where $\hat{\mathbf{b}} = P_2\mathbf{b}, \hat{\mathbf{c}} = P_1\mathbf{c}, \hat{\mathbf{A}} = P_2\mathbf{A}P_1 ,\hat{\circ}=P_2\circ$.
% add the definition for permutation matrix in the appendix 
\end{defn}

We denote that two instances $\mathcal{P}_1$ and $\mathcal{P}_2$ are equivalent or instance-level-isomorphic by $\mathcal{P}_1 \sim \mathcal{P}_2$. Note that we require the model formulation to strictly adhere to the textual description of the problem and account for differences in formulation strength. For example, adding slack variables to the standard instance may not change the optimal solution, but it alters the direct physical meaning specified in the textual description. Therefore, we consider it a different formulation. Consequently, our definition of model equivalence only allows altering the variable notations and permutating variables/constraints, which is in general stricter than that conventionally adopted in OR.

Our concept of model equivalence aligns with the isomorphism of graphs, allowing nodes to be re-indexed or rearranged without changing the graph structure. This motivates us to incorporate tools in graph theory to evaluate model equivalence. Following existing work in learning to optimize \citep{gasse2019exact,chen2022representing}, we represent an LP/MILP model instance by a bipartite graph (see Figure \ref{figure:bipartite_graph} for an illustrative example). We proved that detecting the model equivalence can be reduced to testing graph isomorphism; See Appendix \ref{appendix:equivalence} for a formal demonstration.
%\footnote{See Appendix \ref{appendix:equivalence}}. 
\vspace{-3mm}
\subsection{Model equivalence based on graph isomorphism}
\label{subsection:graph_iso}
\begin{figure}[t]
\begin{center}
%\framebox[4.0in]{$\;$}
%\fbox{\rule[-.5cm]{0cm}{4cm} \rule[-.5cm]{4cm}{0cm}}
\includegraphics[width = 0.5\textwidth]{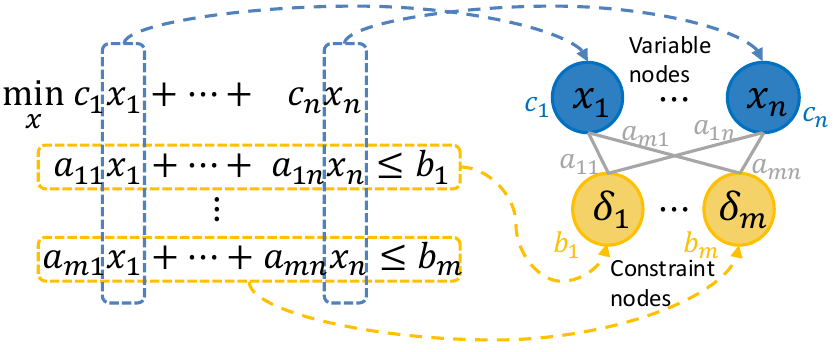}
\end{center}

\caption{Transform model instance to a bipartite graph.}
\label{figure:bipartite_graph}  
\end{figure}

We evaluate the modeling result in two steps:
\vspace{-3mm}
\paragraph{Create test and standard graphs} 
We can use (weighted) bipartite graphs to equivalently represent the modeling problem instance. We follow the formal notation from \cite{chen2022representing} to represent (MI)LP instances to bipartite graphs as follows:

\begin{defn}[Weighted Bipartite Graph Instance Representation]
\label{definition:bipartite_graph}
A MILP/LP instance can be represented as a weighted bipartite graph 
$\mathbf{G} = (\mathbf{V}\cup \mathbf{W},\mathbf{E})$, 
where $\mathbf{V}=\{\mathbf{v}_1,\dots,\mathbf{v}_m\}$ corresponds to constraints 
and $\mathbf{W}=\{\mathbf{w}_1,\dots,\mathbf{w}_n\}$ corresponds to variables. 
Each edge $(v_i,w_j)\in\mathbf{E}$ is weighted by the coefficient $\rmA_{ij}$, 
and each vertex is associated with features (e.g., right-hand side $b_i$, operator type $\circ_i$ for constraints; objective coefficient $c_j$, integrality type $\tau_j$ for variables).
\end{defn}
Since its dependence on $\theta = (\rmA, \rvc, \mathbf{\tau},\rvb, \circ)$, we may write $\rmG(\theta)$; see figure \cref{figure:bipartite_graph} for an example to transform a modeling problem instance to a bipartite graph instance.

As introduced in Definition \ref{definition:bipartite_graph} and Figure \ref{figure:bipartite_graph}, we represent MILP/LP instances as bipartite graphs. In such graphs, nodes can be divided into two groups: variable nodes and constraint nodes. All nodes are equipped with the necessary features. Each constraint node connects with all associated variable nodes. Given such graph representations, we can use graph isomorphism testing tools to detect equivalence between models. 
\vspace{-3mm}
\paragraph{Isomorphism testing} Graph isomorphism testing is a challenging problem, with no known polynomial-time algorithm to date \citep{garey1979computers,babai2016graph}. The Weisfeiler-Lehman test \citep{leman1968reduction} is an effective and computationally efficient approximation for graph isomorphism testing. Typically, one may determine that two graphs are non-isomorphic if the WL-test algorithm produces different outputs (in the form of so-called ``coloring distributions'').  However, if the WL-test yields the same outputs for the two graphs, they are not guaranteed to be isomorphic \citep{cai1992optimal}; See Appendix \ref{appendix: examples} for counterexamples. 

%To prevent misjudgment, we propose two sufficient conditions to make sure that instances satisfying such conditions can be distinguished by the WL-test. In addition, we provide an algorithm to detect whether a given instance satisfies sufficient conditions in Algorithm \ref{alg:sym_decomp}. We applied our sufficient condition detection algorithm to evaluate the problem instances in our benchmark dataset. While we did not intentionally select problem instances based on specific criteria, we found that all problems in Bench4Opt(formally introduced in section \ref{subsection:bench4Opt}) satisfy our sufficient conditions.

In contrast to the widely used WL-test, which offers no theoretical guarantee of producing correct results within polynomial time, our enhanced algorithm first verifies the satisfaction of a sufficient condition. This additional step establishes a formal guarantee of correctness for the subsequent equivalence evaluation.

\vspace{-3mm}
\subsection{Sufficient condition for graph isomorphism testing}
\label{subsection:sufficient_condition}
We propose a sufficient condition, say symmetric decomposable, that modeling instances should satisfy to be testable by a polynomial-time isomorphism testing algorithm.

\vspace{-2mm}
\begin{definition}[Symmetric Decomposable Instance]
\label{definition:decom_sym}
We say a modeling instance $\mathcal{P}$ is \textbf{symmetric decomposable} if, after WL-test, the coloring on its representation graph $\mathcal{G}$ satisfies the following conditions: Excluding nodes that are uniquely colored, the remaining nodes can be divided into $k$ disjoint groups (with some $k\geq 0$) of the same size, denoted by $S_1, S_2, \cdots, S_k$, where

\begin{enumerate}
\vspace{-2mm}
 \item All nodes in the same group have distinct colors,
 % every two node has distinct color xx 
 \vspace{-2mm}
\item All groups share the same coloring sets, and
 \vspace{-2mm}
\item Every two groups are disconnected, i.e. $\forall$ nodes $a\in S_i, b\in S_j, i\neq j$, $a$ is disconnected with $b$.
\end{enumerate}
%consists of several disconnected WL-determinable sub-graphs, and any pair of the sub-graphs are isomorphic.
\end{definition}

\vspace{-2mm}

In previous work, \citet{chen2022representing} characterized one class of graph conditions, termed unfoldable graphs, whose isomorphism can be accurately distinguished by WL-Test. Yet the underlying graphs of many MILP problems do not fall into this category. For example, graphs for bin-packing instances are typically not unfoldable; see example \ref{example:symmetric_example}. The symmetric decomposable condition broadens the scope of problems for which WL-based isomorphism detection is guaranteed. 

While the unfoldable property requires all nodes to have distinct colors, the symmetric decomposable property is more relaxed. It allows the graph to be partitioned into several subgraphs such that within each subgraph, every node has a distinct color. Note that when $k = 1$, a symmetric decomposable problem is reduced to being unfoldable. One example of a decomposable instance can be found in Figure \ref{fig:decomp_symmetric}.

%% check this later 
In the following theorem, we show that if the standard instance is symmetric decomposable, then Algorithm \ref{alg:equal_detection_any} is reliable for detecting whether a test instance is model-equivalent to the standard instance. Rigorous proof can be found in Appendix \ref{appendix:proof_alg_equal}.

\begin{theorem}
\label{theorem:sufficient_equal_detect}
Suppose $\mathcal{P}_{1}$,$\mathcal{P}_{2}$ are symmetric decomposable, then $\mathcal{G}_{1}$ and $\mathcal{G}_{2}$ shares the same coloring distribution after WL-test coloring $\Longleftrightarrow \mathcal{P}_{1}\sim \mathcal{P}_{2}$.
\end{theorem}

To leverage theorem \ref{theorem:sufficient_equal_detect} in practice, we designed an algorithm to identify symmetric decomposable instances (Algorithm \ref{alg:sym_decomp}). Moreover, we prove that under mild assumptions, a random sample yields a symmetric decomposable instance with high probability (Theorem \ref{theorem:thm_sample_continuous}, \ref{theorem:thm_sample_discrete}). Leveraging this property, we construct a benchmark dataset in which all ground-truth instances are guaranteed to be symmetric decomposable.

\vspace{-3mm}
\subsection{ORGEval: Model-Equivalence Evaluation based
 on Graph}
 \vspace{-1mm}
\label{subsection:orgeval}
Combined with the symmetric decomposable detection algorithm (Algorithm \ref{alg:sym_decomp}), we develop ORGEval, a variant form of WL-test to test model equivalence. ORGEval can accurately detect whether a test instance is equivalent to a symmetric decomposable ground truth. It involves 1) running a WL-test for the LP/MILP test and standard instances\footnote{We use the same implementation as \cite{chen2022representing}, presented as Algorithm \ref{alg:WL_MILP}.}; 2) checking whether the test instance is symmetric decomposable based on its coloring distributions (done by Algorithm \ref{alg:sym_decomp}). 
If not, since all ground truth instances are selected to be symmetric decomposable, the test instance must be different from a ground truth one, so the algorithm returns "not equivalent".
If yes, go to step 3. 
3) checking whether the two coloring distributions are identical.
If yes, then return "equivalent"; if not, then return "not equivalent". 

\paragraph{Efficiency of ORGEval}
The time complexity to distinguish tested problem instances from the standard instances with $m$ variables and $n$ nodes is at most $\mathcal{O}(k(m+n)^2)$, where $k$ is the number of clusters in a symmetric decomposable graph. This is far better than the complexity for exhaustive isomorphism testing; detailed complexity analysis can be found in Appendix \ref{subsection: complexity}.

\begin{algorithm}[H]
% sufficient isomorphism testing
\caption{Modeling Equivalence Detection}
\label{alg:equal_detection_any}
\begin{algorithmic}[1]
\small
    \Require Two graph instances $(G_k,H_k) \in \mathcal{G}^k_{m,n}\times \mathcal{H}_m^V\times \mathcal{H}_n^W$ and adjacency matrix $\mathbf{A}_k,k=1,2$; iterate limit $L >0$.
    
    \State Color nodes in two graphs using WL-test Algorithm for MILP/LP.
    \State Get two coloring multi-sets $\mathcal{C}_k=\left\{\{\{C_i^{k,V}\}\}_{i=0}^m,  \{\{C_j^{k,W}\}\}_{j=0}^n\}\right\}, k =1,2$ for coloring $\mathcal{G}_1$ and $\mathcal{G}_2$.
    \State Derive set of unique elements in $\mathcal{C}_k$, denote as set $\mathbb{A}_k, \forall k = 1,2.$
    \If{$\mathcal{C}_1\neq \mathcal{C}_2$}
        \State{\Return Not equivalent}
    \ElsIf{$\mathcal{G}_2$ are symmetric decomposable}
            \State \Return Equivalent
    \Else
    \State\Return Not Equivalent
    \EndIf
    
\end{algorithmic}
\end{algorithm}

% return isomorphism 【check the thm later】
% cite the algorithm outside the algorithm block
% name : augmented WL-test, equivalence detection not universal
% the algorithm Beyond wltest， it is sufficient for our benchmark 【一句话解释】
% 补充一句这个算法是干啥的 -- complexity

%and theoretically prove xxx

%\begin{definition}[WL-Determinable Instance]
%We say a model instance $\mathcal{P}$ is \textbf{WL-determinable} if WL-test of its underlying graph outputs distinct colors for different nodes in its graph representation.
%\label{definition:wl_det}
%\end{definition}

%This definition aligns with the definition of unfoldable graphs in \cite{chen2022representing}.

%\input{content/5_dataset}
\vspace{-3mm}
\section{Experiment and Analysis}
\label{section:exp}
\vspace{-3mm}
We present comprehensive experiments on ORGEval using Bench4Opt, the first dataset for optimization modeling that separates models from data. Bench4Opt comprises 209 LP and MILP instances curated from both hand-crafted problems and the MIPLIB dataset \citep{miplib}. We empirically demonstrate the efficiency and consistency of ORGEval in Section \ref{subsection:exp_advantage}, and further employ it to benchmark the performance of LLMs in optimization modeling in Section \ref{subsection:exp_benchmark}.

\vspace{-3mm}
\subsection{Benchmark dataset}
\vspace{-1mm}
\label{subsection:bench4Opt}
To robustly assess ORGEval, we introduce Bench4Opt, a diverse benchmark dataset consisting of 394 optimization modeling word problems in a model-data separated format, each with two levels of abstract, structured, and unstructured description. Specifically, each Bench4Opt problem instance comprises three components: 1) problem description, 2) parameter file, and 3) reference model in .lp format.

An illustrative example from Bench4Opt can be found in Figure \ref{fig:wp}. Besides careful verification and quality control, we applied our sufficient condition detection algorithm \cref{alg:sym_decomp} to evaluate the problem instances in our benchmark dataset. While we did not intentionally select problem instances based on specific criteria, we found that all problems in Bench4Opt satisfy our sufficient conditions.

\vspace{-3mm}
\subsection{Advantages of ORGEval}
\vspace{-2mm}
\label{subsection:exp_advantage}
\paragraph{Evaluation Efficiency}
Empirically, we demonstrate that ORGEval offers significantly higher evaluation efficiency compared with solver-based methods, especially for problem instances that are challenging for solvers. To test this, we sampled 75 problem instances from the MIPLIB \citep{miplib} dataset across three difficulty levels: easy, hard, and open for existing solvers, with 25 instances per level. According to MIPLIB's definition, easy instances can be solved within an hour using a standard solver on a typical desktop machine with up to 16 threads; hard instances require a longer time, non-standard hardware, or advanced algorithms; and open instances have not yet been solved. Solver often requires hours or more to evaluate selected instances from MIPLIB. In contrast, ORGEval consistently produced evaluation results within a reasonable timeframe--- ORGEval runs 30 seconds on average to output an evaluation result, even for the most challenging open problems for solvers. See Table~\ref{tab:efficiency_miplib} for the average evaluation time of ORGEval across the three difficulty levels.

\vspace{-3mm}
\begin{table}[H]
\caption{Evaluation time of ORGEval for three levels of difficulties: easy, hard, open. Instances are sampled from MIPLIB, with 25 instances per level.}
\label{tab:efficiency_miplib}
\resizebox{\textwidth}{!}{
\begin{tabular}{l|ccc}
\toprule
\textbf{Level of Difficulty} & \textbf{\begin{tabular}[c]{@{}c@{}}Average Problem Size\\  (\#variables + \# constraints)\end{tabular}} & \textbf{\begin{tabular}[c]{@{}c@{}}Average Evaluation Time \\ (Solver)\end{tabular}} & \textbf{\begin{tabular}[c]{@{}c@{}}Average Evaluation Time \\ (ORGEval)\end{tabular}} \\ \hline
Easy                         & 1848                                                                                                    & about 1 hour                                                                       & \textbf{0.21s}                                                                        \\
Hard                         & 10463                                                                                                   & $> $ 1 hour                                                                    & \textbf{3.83s}                                                                        \\
Open                         & 17050                                                                                                   & not yet being solved                                                                 & \textbf{32.07s}                                                                       \\ \bottomrule
\end{tabular}
}
\end{table}
\vspace{-3mm}

\paragraph{Evaluation Consistency}
We use the evaluation result of five random instances to indicate equivalence between two models. For such an evaluation to be valid, we address the consistency of our evaluation result across various data configurations: Our experimental results show that ORGEval achieves 100\% consistency across five random data configurations for all models in Bench4Opt. In contrast, solvers fail to evaluate models with random data configurations for more than 60\% of the models due to infeasibility, and even among the remaining models, 5.89\% of model pairs yield inconsistent results under solver-based equivalence evaluation.

\vspace{-3mm}
\begin{table}[H]
\centering
\caption{Comparison of model consistency across 5 random instances under different evaluation schemes.}
\resizebox{0.9\textwidth}{!}{%
\begin{tabular}{lccccl}
\toprule
\textbf{Model} & \textbf{Feasibility consistency} & \textbf{ORGEval consistency} & \textbf{Solver consistency} \\
\midrule
gpt-4o                        & 36.30\% & 100.00\% & 95.58\%\\
claude-opus-4       & 36.05\% & 100.00\% & 92.12\% \\
deepseek-v3                & 34.69\% & 100.00\% & 93.95\% \\
o1                   & 35.43\% & 100.00\% & 94.77\%  \\
\midrule
\textbf{Average}             & 35.62\% & 100.00\% & 94.11\% \\
\bottomrule
\end{tabular}
}
\label{tab:instance-consistency}
\end{table}
\vspace{-5mm}
\subsection{Benchmark the modeling ability using ORGEval and Bench4Opt}
\label{subsection:exp_benchmark}
To assess the capabilities of LLMs in optimization modeling, we conducted a comprehensive evaluation using the Bench4Opt benchmark. Our evaluation focused on top-performing LLMs via direct prompting. The main evaluation result is listed in Table \ref{tab:main_result}.

% Please add the following required packages to your document preamble:
% \usepackage{booktabs}
% \usepackage{multirow}
% \usepackage{graphicx}
% \usepackage[table,xcdraw]{xcolor}
% Beamer presentation requires \usepackage{colortbl} instead of \usepackage[table,xcdraw]{xcolor}
\begin{table}[H]
\begin{center}

\vspace{-1mm}
\caption{Evaluation Results on Bench4Opt. SOTA in each category is marked in red.}
\label{tab:main_result}
\resizebox{0.75\textwidth}{!}{%
\begin{tabular}{@{}lcccc@{}}

\toprule

\multicolumn{1}{l|}{}    
    & \multicolumn{2}{c|}{\textbf{Overall}} 
	& \multicolumn{2}{c}{\textbf{Modeling Accuracy}}  
\\ \cmidrule(l){2-5} 

\multicolumn{1}{l|}{\multirow{-3}{*}{\textbf{LLMs}}} 
    & \multicolumn{1}{c|}{\textbf{Accuracy}} 
	& \multicolumn{1}{c|}{\textbf{Compile Error}} 
	& \multicolumn{1}{c|}{\textbf{Structured}} 
	& \multicolumn{1}{c}{\textbf{Unstructured}}

\\ \midrule                                                         
\multicolumn{1}{l|}{deepseek-v3}
& {\color[HTML]{CB0000}54.82} 
& \multicolumn{1}{c|}{2.28}                      
& {\color[HTML]{CB0000}63.45}
& \multicolumn{1}{c}{46.19}\\

\multicolumn{1}{l|}{claude-opus-4}
& {\color[HTML]{CB0000}54.82}    
& \multicolumn{1}{c|}{2.28}                         
&  60.41
& \multicolumn{1}{c}{\color[HTML]{CB0000}49.24}\\

\multicolumn{1}{l|}{gpt-4.1}                                           
& 52.28
& \multicolumn{1}{c|}{1.78}                   
& 57.36                                    
& \multicolumn{1}{c}{47.21}\\

\multicolumn{1}{l|}{gpt-4o}                                            
& 51.02   
& \multicolumn{1}{c|}{7.36}               
& 58.38                                  
& \multicolumn{1}{c}{43.65}\\

\multicolumn{1}{l|}{claude-opus-4.1} 
& 50.76            
& \multicolumn{1}{c|}{\color[HTML]{CB0000}0.76}
&  59.39                           
& \multicolumn{1}{c}{42.13}\\

\multicolumn{1}{l|}{o3}  
& 47.97       
& \multicolumn{1}{c|}{\color[HTML]{CB0000}0.76}
& 55.84                                
& \multicolumn{1}{c}{40.10}\\

\multicolumn{1}{l|}{deepseek-r1}  
 & 47.72   
& \multicolumn{1}{c|}{2.28}
& 55.84                               
& \multicolumn{1}{c}{39.59}                                           \\

\multicolumn{1}{l|}{o1}  
 & 47.21   
& \multicolumn{1}{c|}{1.78}
& 52.79                                 
& \multicolumn{1}{c}{41.62}                                           \\

\bottomrule
\end{tabular}%
}
\end{center}
\end{table}

\vspace{-5mm}

Our benchmark revealed varying performances among the tested models. In particular, claude-opus-4 \citep{claude4} and deepseek-v3 \citep{deepseek} achieved the strongest results, each reaching an overall accuracy of 54.82\%, outperforming other contenders across structured and unstructured optimization tasks. In contrast, reasoning models such as deepseek-r1 \citep{deepseek}, o1 \citep{o1openai}, and o3 \citep{o3openai} consistently exhibited lower accuracy compared to the base model. While reasoning models produce outputs with relatively fewer compile errors, their multi-step reasoning capabilities appear susceptible to hallucination propagation. The cascading effect of these reasoning artifacts likely contributes to progressive accuracy degradation throughout complex problem-solving sequences. This phenomenon may explain the observed performance gap despite their enhanced error-handling capabilities.
\vspace{-2mm}
\section{Conclusion}
\vspace{-2mm}
In this work, we formalize the task of detecting equivalence between (MI)LP models and introduced a new modeling equivalence evaluation method, \textbf{ORGEval}, to evaluate the equivalence between optimization models. By representing optimization models as graphs and leveraging the Weisfeiler-Lehman (WL) test under well-defined sufficient conditions, our method offers a principled and efficient alternative to solver-based evaluations. Our experiments demonstrate that ORGEval achieves consistent evaluation results across all data configurations and offers significant speed advantages over solver-based methods, particularly on hard-to-solve problems. To access ORGEval , we also introduce Bench4Opt, a diverse benchmark dataset of 394 model-data-separated optimization problems containing problem from MIPLIB dataset and hand crafted problem generated with the help of LLM.

%\subsubsection*{Author Contributions}
%If you'd like to, you may include  a section for author contributions as is done
%in many journals. This is optional and at the discretion of the authors.

%\subsubsection*{Acknowledgments}
%Use unnumbered third level headings for the acknowledgments. All
%acknowledgments, including those to funding agencies, go at the end of the paper.

\bibliography{main}
\bibliographystyle{iclr2026_conference}

\newpage
\appendix
% \section{LLM Usage Statement}
% In this work, we used large language models (LLMs) in two ways:

% \paragraph{Paper Revision:} LLMs were used to assist in improving the clarity and readability of the manuscript. All scientific content, results, and interpretations were reviewed, verified, and rewritten by the authors.

% \paragraph{Benchmark Construction:} LLMs played a role in constructing and expanding the benchmark dataset. Specifically, they assisted in generating candidate problem instances and verifying problem formulations. All final benchmark data and evaluations were curated and validated by the authors to ensure correctness and consistency.

\section{Data Construction}
\label{appendix: data_construction}
\subsection{Dataset}
% Please add the following required packages to your document preamble:
% \usepackage{booktabs}
% \usepackage{multirow}
% \usepackage{graphicx}
% Please add the following required packages to your document preamble:
% \usepackage{booktabs}
% \usepackage{multirow}
% \usepackage{graphicx}
\begin{table}[H]
\begin{center}

\caption{Optimization problem types and classes including in our Bench4Opt.}
\label{tab:type_class}
% \resizebox{0.6\textwidth}{!}{%
\begin{tabular}{@{}c|l@{}}
\toprule
\multicolumn{1}{l|}{Problem Types} & Problem Classes                       \\ \midrule
\multirow{8}{*}{LPs}               & Diet Problem                          \\
                                   & Transportation Problem                \\
                                   & Blending Problem                      \\
                                   & Production Planning Problem           \\
                                   & Network Flow Problem                  \\
                                   & Portfolio Optimization Problem        \\
                                   & Cutting Stock Problem                 \\
                                   & Staff Scheduling Problem'             \\ \midrule
\multirow{8}{*}{MILPs}             & Knapsack Problem                      \\
                                   & Traveling Salesman Problem (TSP)      \\
                                   & Vehicle Routing Problem (VRP)         \\
                                   & Bin Packing Problem                   \\
                                   & Set Covering Problem                  \\
                                   & Capacitated Facility Location Problem \\
                                   & Capital Budgeting Problem             \\
                                   & Assignment Problem                    \\ \bottomrule
\end{tabular}%
% }
\end{center}
\end{table}
\begin{figure}[H]
\begin{center}
%\framebox[4.0in]{$\;$}
%\fbox{\rule[-.5cm]{0cm}{4cm} \rule[-.5cm]{4cm}{0cm}}
\includegraphics[width=0.9\textwidth]{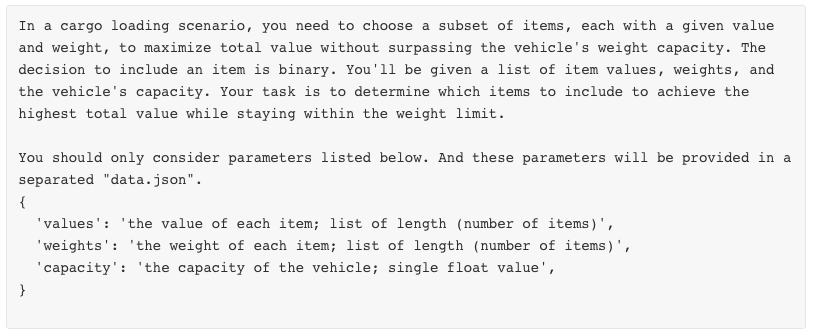}
\end{center}
\caption{Example for concise version word problem on cargo loading.}
\label{fig:wp_concise}
\end{figure}

\begin{figure}[H]
\begin{center}
%\framebox[4.0in]{$\;$}
%\fbox{\rule[-.5cm]{0cm}{4cm} \rule[-.5cm]{4cm}{0cm}}
\includegraphics[width=0.9\textwidth]{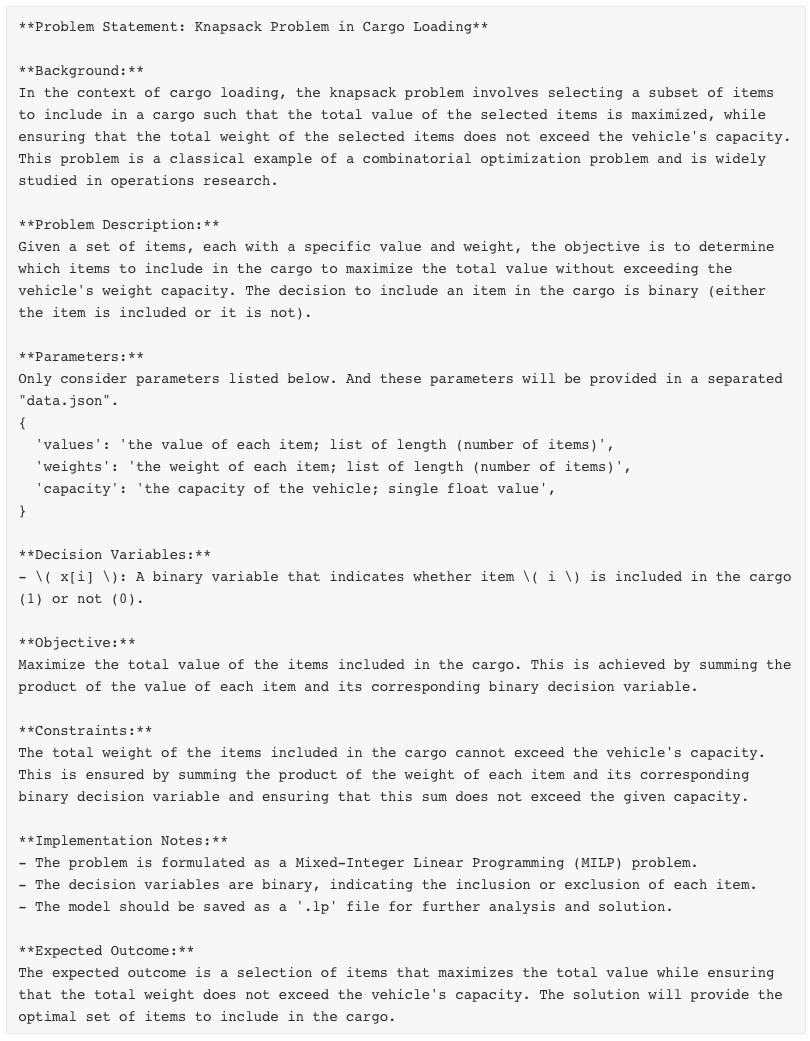}
\end{center}
\caption{Example for word problem on cargo loading.}
\label{fig:wp}
\end{figure}

\begin{figure}[H]
\begin{center}
%\framebox[4.0in]{$\;$}
%\fbox{\rule[-.5cm]{0cm}{4cm} \rule[-.5cm]{4cm}{0cm}}
\includegraphics[width=0.6\textwidth]{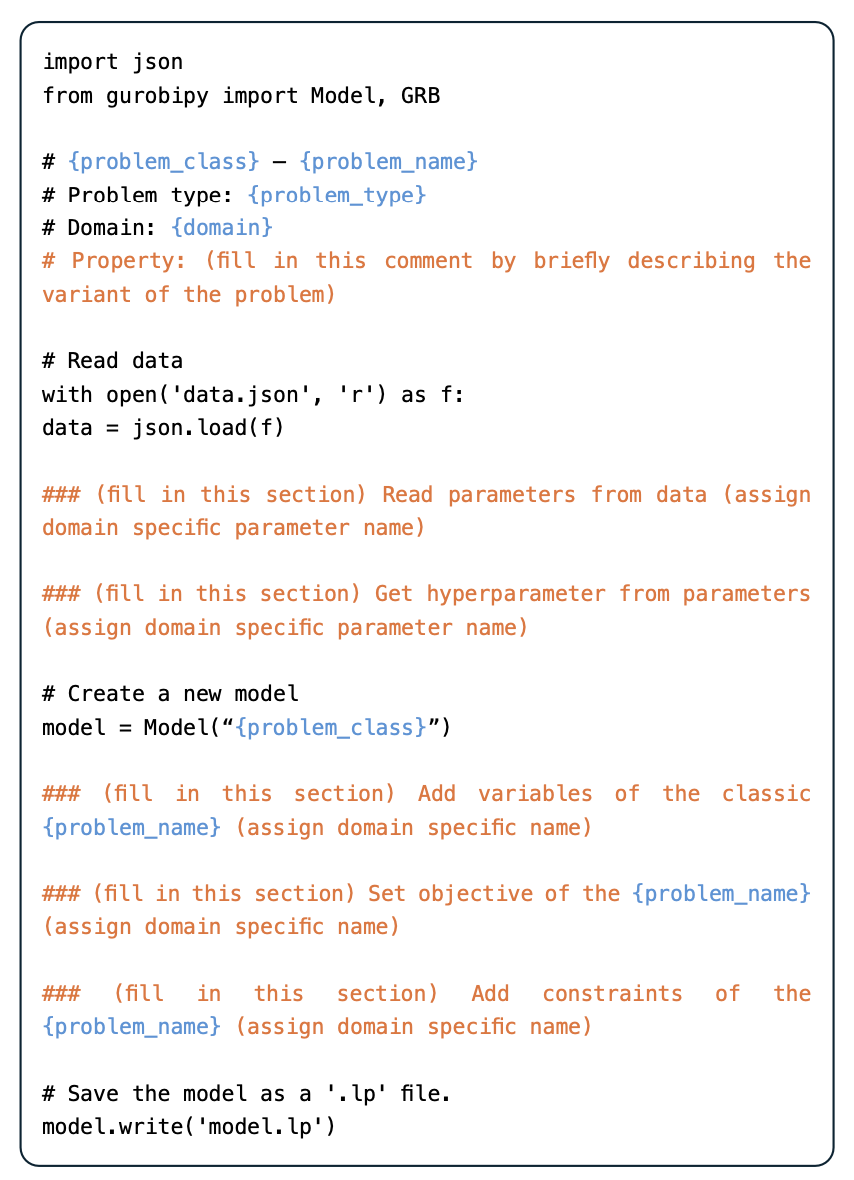}
\end{center}
\caption{Code skeleton for optimization model simulation.}
\label{fig:code_skeleton}
\end{figure}

\begin{figure}[H]
\begin{center}
%\framebox[4.0in]{$\;$}
%\fbox{\rule[-.5cm]{0cm}{4cm} \rule[-.5cm]{4cm}{0cm}}
\includegraphics[width=0.6\textwidth]{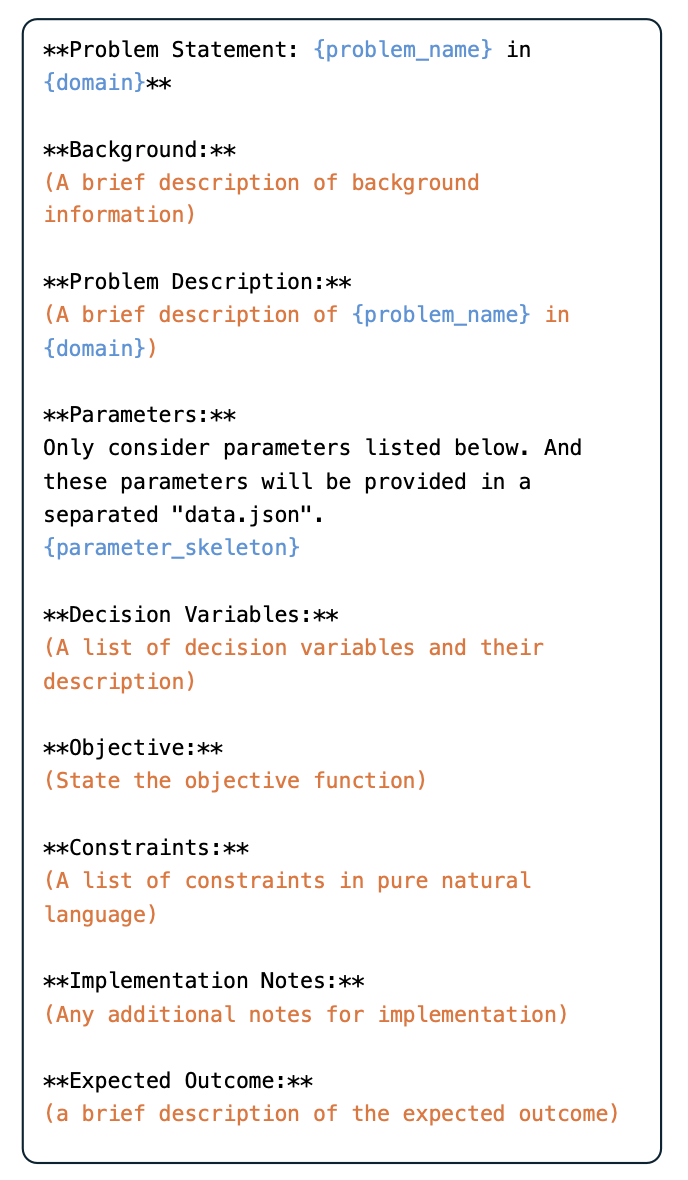}
\end{center}
\caption{Standard structure for word problem crafted from INFORMS AIMMS-MOPTA Optimization Modeling Competition.}
\label{fig:prompt_skeleton}
\end{figure}

\newpage

\section{Related work}
\paragraph{NLP for OR Modeling}
While substantial progress has been made in automatic modeling of general mathematical problems \citep{bobrow1964question, dellarosa1986computer, sundaram2015natural, liu2025mm}, there has been limited focus on applying these techniques specifically to operations research. Prior to the rise of LLMs, the NL4Opt competition \citep{pmlr-v220-ramamonjison23a} explored the feasibility of learning-based natural language interfaces for optimization solvers. More recently, works leveraging LLMs, such as the Chain-of-Experts (CoE) \citep{xiao2023chain} and OptiMUS \citep{ahmaditeshnizi2024optimus}, introduced multi-agent cooperative systems to model and code complex OR problems automatically. Furthermore, \cite{tang2024orlm} fine-tuned open-source LLMs with approximately 7B parameters, achieving significant performance improvements over baseline models. These advancements underscore the immense potential of LLMs in optimization modeling. With the emergence of LLMs, there is an increasing need for benchmarks to understand their capability boundaries \citep{liu2024mathbench, zhou2024your, sawada2023arb}.
Several optimization modeling benchmarks have been proposed to evaluate LLMs. The Linear Programming Word Problem (LPWP) dataset \citep{ramamonjison2022augmenting} includes multiple domains and comprises up to 1,001 LP problems. However, it primarily consists of elementary-level LP problems.
%, limiting its effectiveness in assessing advanced modeling capabilities. 
The ComplexOR dataset \citep{xiao2023chain} was designed to feature more complicated OR problems, but its limited size and inclusion of numerical data within the textual description still constrain the level of complexity it can represent. IndustryOR \citep{tang2024orlm}, MAMO \citep{huang2024mamo}, and E-OPT \citep{yang2024benchmarking} strive to cover a broader range of OR problems through data synthesis and augmentation. The NLP4LP dataset \citep{ahmaditeshnizi2024optimus} attempts to separate data from textual descriptions, yet the problem sizes remain small, and the descriptions are well structured with variables/constraints/objectives explicitly presented. In comparison to existing benchmarks, our work aims to provide a more comprehensive dataset and a more rigorous evaluation method, enabling a more precise assessment of LLM capabilities in optimization modeling.

\paragraph{Modeling Equivalence Evaluation} The earliest work to evaluate NLP for OR modeling performance is to calculate the canonical accuracy \citep{ramamonjison2022augmenting}. This accuracy counts for the declaration-level(e.g., objective or constraints) matching score between predicted and reference formulations. This method has severe limitations as it's highly sensitive to superficial differences in formulation, such as variable naming or ordering. More recent benchmark works—including MAMO \citep{huang2024mamo}, IndustryOR \citep{tang2024orlm}, NLP4LP \citep{ahmaditeshnizi2024optimus}, and OptiBench \citep{yang2024optibench}, and equivamap \citep{zhai2025equivamap}—relies on solvers to assess modeling quality. They execute the predicted numerical models and compare the resulting optimal values with reference optimal values to evaluate correctness. While solver-based approaches better align with the functional goals of optimization, they introduce new limitations. The evaluation becomes dependent on solver behavior, which is often unstable, especially when the focus is on model structure equivalence rather than model instance outcome equivalence. For instance, small changes in parameters can render a model infeasible or non-convex, causing solvers to fail or return suboptimal solutions. As a result, optimal value mismatches may stem not from modeling errors but from solver or numerical issues, thereby confounding the reliability of equivalence assessment.

\paragraph{Broader Research on AI for OR}
Beyond model formulation, significant progress has been made in the field of AI for Operations Research (AI for OR), particularly in parameter generation and solving optimization problems \citep{rajgopal2004principles}. In parameter generation, AI techniques have been employed for better simulation of key parameters of optimization problems \citep{elmachtoub2022smart, maragno2023mixed, bergman2022janos}. Similarly, our work leverages LLMs to generate necessary problem data through a program of thoughts \citep{chen2022program}. On the optimization side, numerous studies have focused on leveraging AI models in automatic algorithm configuration \citep{ansotegui2009gender,lindauer2022smac3,anastacio2020combining}, optimization algorithm selection \citep{wang2019satnet,chi2022deep}, and heuristic algorithm design \citep{zeng2022reinforcement,talbi2009metaheuristics,romera2024mathematical}. Specifically, a line of research has modeled MILP/LP problems as bipartite graphs and applied Graph Neural Networks (GNNs) to make decisions at various stages of their solution processes \citep{gasse2019exact, zhou2020graph}. These GNN-based methods have demonstrated efficacy in tasks such as variable selection and node branching, leading to significant improvements in solver performance. Inspired by this, we model optimization problems as bipartite graphs and formalize the evaluation paradigm based on the classical WL-test algorithm \cite{leman1968reduction}.

\section{Equivalence Evaluation}
\subsection{Model Equivalence Class}
\label{appendix:model_equivalence}
\begin{definition}[Model Equivalence]

We say $\mathcal{C}(\mathcal{P})$ is a \textbf{model equivalence class} of the MILP/LP problem instance $\mathcal{P}$ if $\forall \hat{\mathcal{P}} \in \mathcal{C}(\mathcal{P}), \exists$ permutation matrices $P_1, P_2$ which shuffles the index of a vector or column index of a matrix s.t. $\hat{\mathcal{P}}$ can be written in the following form:

\begin{align*}
\min_x& \hat{c}^Tx, \\
\text{s.t. }& \hat{A} x \hat{\circ} \hat{b}, \hat{l}\leq x\leq \hat{u}
\end{align*}

where $\hat b = P_2b, \hat C = P_1C, \hat A = P_2AP_1 ,\,\hat\circ=P_2\circ,\,\hat l=P_1l, u=P_1u$.
% add the definition for permutation matrix in the appendix 

$\forall \mathcal{P}_2 \in \mathcal{C}(\mathcal{P}_1)$, we say $\mathcal{P}_2$ is \textbf{model-equivalent} to $\mathcal{P}_1$, denote as $\mathcal{P}_1 \sim \mathcal{P}_2$.
\label{definition:model_equivalent}
\end{definition}

\subsection{Weighted Bipartite Graph for Representing MILP/LP}
\label{appendix:bipartite}
A weighted bipartite graph for a MILP/LP instance is denoted by $\mathbf{G} = (\mathbf{V}\cup \mathbf{W},\mathbf{E})$, with vertex set $\mathbf{V}\cup \mathbf{W}$ divided into 2 groups $\mathbf{V}=\{\mathbf{v}_1,\cdots,\mathbf{v}_m\}$ for constraints, and $\mathbf{W}=\{\mathbf{w}_1,\cdots,\mathbf{w}_n\}$ for variables, $\mathbf{E}$ consisting of $E_{ij}= E(v_i,w_j)$, 
$\forall i = 1,\cdots,m, j = 1,\cdots,n$.
To fully represent all information in a MILP/LP instance, we associate each vertex with features:
\begin{itemize}
\item The constraint vertex $\mathbf{v}_i \in \mathbf{V}$ is equipped with a feature vector $\mathbf{H}^V$ with elements $\mathbf{h}_{i}^V = (b_i,o_i)\in \mathcal{H}^V = \mathbb{R}\times\{\leq,\ge,=,<,>\}$
\item The variable vertex $\mathbf{w}_j \in \mathbf{W}$ is equipped with a feature vector $\mathbf{H}^W$ with elements $\mathbf{h}_{j}^W = (c_j,\tau_j)\in \mathcal{H}^W = \mathbb{R}\times\{\mathbb{R}\cup-\infty\}\times\{\mathbb{R}\cup\infty\}\times\{0,1\}.$ $\tau_j = 1$ if $j \in \mathbb{Z}$ and $\tau_j = 0$ otherwise.
\end{itemize}
The edge $E_{ij} \in \mathbb{R}$ connects $\mathbf{v}_i \in \mathbf{V}$ and $\mathbf{w}_j \in \mathbf{W}$, $E_{ij} = \mathbf{A}_{ij}$. There is no edge connecting vertices in the same vertex group.

\subsection{Connection between Model Equivalence and Graph Isomorphism} 
To test whether 2 modeling instances were permutation equivalent, we can equivalently conduct isomorphism testing between their corresponding weighted bipartite graphs. Lemma \ref{lemma:lemma1} establishes an equivalence between assessing modeling appropriateness and graph isomorphism testing.
\label{appendix:equivalence}
\begin{definition}[Graph Isomorphism]
\label{definition: def_iso}
Consider 2 graphs $\mathcal{G}_1 = (\mathbf{G}_1,\mathbf{H}^V_1\times \mathbf{H}^W_1)$ and $\mathcal{G}_2 = (\mathbf{G}_2,\mathbf{H}^V_2\times\mathbf{H}^W_2)$ with $\mathbf{G}_i=(\mathbf{V}^i\cup \mathbf{W}^i,\mathbf{E}^i)|_{1\le i\le 2}$. We say $\mathcal{G}_1$ and $\mathcal{G}_1$ are \textbf{isomorphic} if there exists permutation matrix $\mathbf{P}_1, \mathbf{P}_2$such that:
$\mathbf{P}_1\mathbf{E}^1P_2^T=\mathbf{E}^2, \mathbf{P}_1\mathbf{H}_1^W = \mathbf{H}_2^W, \mathbf{P}_2\mathbf{H}_1^V = \mathbf{H}_2^V$.
If 2 graphs $\mathcal{G}_1$ and $\mathcal{G}_1$ are isomorphic, denote $\mathcal{G}_1 \stackrel{g}{\sim} \mathcal{G}_2$.
\end{definition}

\begin{lemma}
$\forall$ MILP/LP instances $\mathcal{P}_1, \mathcal{P}_2$ with corresponding bipartite graph $\mathcal{G}_1, \mathcal{G}_1$, we have
$$
\mathcal{P}_1 \sim \mathcal{P}_2 \Longleftrightarrow \mathcal{G}_1 \stackrel{g}{\sim} \mathcal{G}_2.
$$
\label{lemma:lemma1}
\end{lemma}
\subsection{Proof of lemma \ref{lemma:lemma1}:} 
We prove this lemma by proving 2 claims:
\paragraph{Claim 1:} $\mathcal{G}_1\sim \mathcal{G}_2 \Longrightarrow \mathcal{P}_1 \sim \mathcal{P}_2$.

Suppose $\mathcal{G}_1\sim \mathcal{G}_2$. For bipartite graphs $\mathcal{G}_1$ and $\mathcal{G}_2$, nodes $v_i$ would only connect with some node $w_j$ if the $j$-th constraint involves decision variable $x_i$. Therefore the adjacency matrix of $\mathcal{G}_k$ would be in the form $\mathbf{A^{(k)}_{adj}} = \left[ \begin{matrix}0 & \mathbf{A}_k^T\\ \mathbf{A}_k &0 \end{matrix}\right], \forall k =1,2.$
Now, by the assumption that $\mathcal{G}_1\sim \mathcal{G}_2$, $\exists$ permutation matrix $\mathbf{P}$ such that 
$$
\begin{aligned}
&\mathbf{P} = \left[\begin{matrix} \mathbf{P}_V & 0\\ 0& \mathbf{P}_W\end{matrix}\right],\\
&\mathbf{P}\mathbf{A}^{(1)}_{adj}\mathbf{P} = \mathbf{A}^{(2)}_{adj},\\
&\mathbf{P}_V^T\mathbf{H}^V_1 = \mathbf{P}^T\mathbf{H}^V_2,\\
&\mathbf{P}_W^T\mathbf{H}^W_1 = \mathbf{P}_W^T\mathbf{H}^W_2.
\end{aligned}
$$
Therefore, we have
$$
\mathbf{A}^{(2)}_{adj} = \begin{bmatrix}0&\mathbf{P}_V\mathbf{A}_1^T\\\mathbf{P}_W\mathbf{A}_1&0\end{bmatrix} \text{ and } \mathbf{H}_2 = \begin{bmatrix}\mathbf{P}_V\mathbf{H}^V_1\\\mathbf{P}_W\mathbf{H}^W_1 \end{bmatrix}.
$$
We may reformulate the MILP/LP instance $\mathcal{P}_2$ as follows:
$$
\begin{aligned}
\mathcal{P}_2: \quad &\min_{\mathbf{x}\in \mathbb{R}^p\times \{0,1\}^{n-p}} \mathbf{c}^T\mathbf{P}_V\mathbf{x}, \\
&\text{s.t. } \mathbf{P}_W\mathbf{A}\mathbf{P}_V \mathbf{x} \circ \mathbf{P}_W\mathbf{b}, \mathbf{l}\leq \mathbf{P}_V\mathbf{x}\leq \mathbf{u},
\end{aligned}
$$
By the definition of permutation equivalent, we say $\mathcal{P}_2 \sim \mathcal{P}_1$.
\paragraph{Claim 2:} $\mathcal{P}_1 \sim \mathcal{P}_2 \Longrightarrow \mathcal{G}_1\sim \mathcal{G}_2$.

Suppose $\mathcal{P}_1 \sim \mathcal{P}_2$. By the definition of permutation equivalent class, $\exists$ permutation matrix $\mathbf{P}_1$ and $\mathbf{P}_2$ such that 
$$
\begin{aligned}
&\mathbf{A}_2 = \mathbf{P}_2\mathbf{A}_1\mathbf{P}_1\\
&\mathbf{b}_2 = \mathbf{P}_2\mathbf{b}_1, \\
&\mathbf{C}_2 = \mathbf{P}_1^T\mathbf{C}_1,\\
&\mathbf{P}_2\circ_1 = \circ_2,
\end{aligned}
$$
Therefore, the corresponding adjacent matrix in the bipartite graph of $\mathcal{P}_2$ is 
$$
\begin{aligned}
\mathbf{A}^{(2)}_{adj} &=\begin{bmatrix}0&\mathbf{A}_2^T\\\mathbf{A}_2&0\end{bmatrix}\\
&=\begin{bmatrix}0&\mathbf{P}_1^T\mathbf{A}_1^T\mathbf{P}_2^T\\\mathbf{P}_2\mathbf{A}_1\mathbf{P}_2&0\end{bmatrix}\\
&= \begin{bmatrix}\mathbf{P}_1^T&0\\0&\mathbf{P}_2\end{bmatrix}\begin{bmatrix}0&\mathbf{A}_1^T\\\mathbf{A}_1&0\end{bmatrix}\begin{bmatrix}\mathbf{P}_1&0\\0&\mathbf{P}_2^T\end{bmatrix}\\
&=\hat{\mathbf{P}}^T\mathbf{A}^{(1)}_{adj} \hat{\mathbf{P}}
\end{aligned}
$$
In addition, we have $\mathbf{b}_2 = \mathbf{P}_2\mathbf{b}_1, \mathbf{c}_2 = \mathbf{P}_1^T\mathbf{c}_1$. Therefore,
$$
\mathbf{H}_2 = \begin{bmatrix}\mathbf{H}_2^V\\\mathbf{H}_2^W \end{bmatrix}=  \begin{bmatrix}\mathbf{P}_1^T&0\\0&\mathbf{P}_2\end{bmatrix}\begin{bmatrix}\mathbf{H}_1^V\\\mathbf{H}_1^W\end{bmatrix} = \hat{\mathbf{P}}^T\mathbf{H}_1.
$$
According to the definition of graph isomorphism, $\mathcal{G}_1$ is isomorphic to $\mathcal{G}_2$.

\subsection{Algorithms}
\label{subsection:algorithms}
\begin{minipage}{\textwidth}
%\paragraph{Weisfei Lehman Test for MILP/LP}
\begin{algorithm}[H]
\caption{WL test for MILP/LP Graphs}
\label{alg:WL_MILP}
\begin{algorithmic}[1]
\Require A graph instance $(G,H) \in \mathcal{G}_{m,n}\times \mathcal{H}_m^V\times \mathcal{H}_n^W$ and iterate limit $L >0$.
\State Initialize with $C_i^{0,V} = HASH_{0,V}(h_i^V)$,  $C_j^{0,W} = HASH_{0,W}(h_j^W)$
\For{$l = 1,2,\cdots,L$}
\State \quad $C_i^{l,V} = HASH(C_i^{l-1,V},\sum_{j = 1}^n E_{i,j}HASH'_{l,W}(C_j)^{l-1,W})$
\State \quad $C_i^{l,W} = HASH(C_i^{l-1,W},\sum_{j = 1}^n E_{i,j}HASH'_{l,V}
(C_j)^{l-1,V})$
\EndFor
\State\Return The multisets containing all colors $\{\{C_i^{L,V}\}\}_{i=0}^m, \{\{C_i^{L,W}\}\}_{j=0}^n$.
\end{algorithmic}
\end{algorithm}

\begin{algorithm}[H]
\caption{Determine if the graph is symmetric decomposable}
\label{alg:sym_decomp}
\begin{algorithmic}[1]
\Require Graph $\mathcal{G}$'s adjacent matrix $\textbf{A}$ and type 2 stable partition sets of it's variable nodes $\mathcal{I} = \left\{I_1,I_2,\cdots,I_{s'} \right\}$ and constraint nodes $\mathcal{J} = \left\{J_1,J_2,\cdots,J_{t'} \right\}$.
\Ensure Returns \textbf{True} if the graph is decomposable symmetric; otherwise, \textbf{False}.
\State $k \gets |I_1|$.
\If{$|I_s|\neq k$ or $|J_t|\neq k$ for some $s = 1,\cdots,s', t= 1,\cdots,t'$.}
    \State\Return \textbf{False}
\Else
    \State Initialize an empty Cluster dictionary $Cluster$
    \For{$i\gets 0$ to $k-1$}
        \State $C \gets$ the set of all numbers for type 2 stable partition sets
        \State Initialize an empty cluster set $Cluster[i]$, initialize an empty queue $Q$.
        \While{Set $C$ is not empty}
            \If{Q is empty}
                \State Randomly select a color $c\in C$, delete $c$ from $C$.
                \State $P_c \gets$ the list of nodes labeled with $c\in C$.
                \State $Cluster[i] \gets [P_c[i]]$, delete node $P_c[i]$ from $S$, push $P_c[i]$ in $Q$.
            \Else
                \While{Q not empty}
                    \State $u \gets Q.dequeue()$
                    \For{neighborhood node $w$ of $u$}
                        \If{w is not in any of $P_c$ or $w$ is in $Cluster[i]$}
                            \State continue
                        \ElsIf{$color(w)$ appears in $Cluster[i]$}
                            \State \Return \textbf{False}
                        \Else
                            \State Add $w$ in $Cluster[i]$, delete color($w$) from $C$, push w in Q.
                        \EndIf
                    \EndFor
                \EndWhile
            \EndIf
        \EndWhile
    \EndFor
    \If{colors in $Cluster[i] \neq 1$ are not distinct for some $i=0,\cdots k-1$} \Comment{check distinct color}
        \State \Return \textbf{False}
    \ElsIf{checkDisjointness($[S^1,\cdots,S^{k-1}]$)} \Comment{check disjointness}
        \State\Return \textbf{False}

    \ElsIf{checkConnectivity($[S^1,\cdots,S^{k-1}]$)} \Comment{check disconnectivity}
        \State \Return \textbf{False}
    \EndIf
\EndIf
\State \Return True
\end{algorithmic}
\end{algorithm}
Notation: We denote the collection of all nodes $v_i's$ indexed by $i\in I_p$ as $\mathbf{I}_p$. 

Function \texttt{checkDisjointness}($[Cluster[1],\dots,Cluster[k-1]]$) outputs \textit{True} 
if any two sets $Cluster[i]$ and $Cluster[j]$, where $i \neq j$, share a common element.

Function checkConnectivity($[Cluster[1],\cdots,Cluster[k-1]]$) output True if there exists some nodes $s \in Cluster[i],\allowbreak s'\in Cluster[j], i\neq j$ such that $s$ connected with $s'$.
\end{minipage}

%\paragraph{Symmetric Detection Algorithm}

\subsection{Proof Preparation for Theorem \ref{theorem:sufficient_equal_detect}}
\label{appendix:proof_alg_equal}
%\begin{theorem}
%Denote Algorithm \ref{alg:equal_detection} by $\mathcal{A}(\mathcal{G}_{test},\mathcal{G}_{standard})$. Suppose $\mathcal{P}_{standard}$ is  WL-determinable or decomposable symmetric, then $\forall \mathcal{P}_{test}$, we have $\mathcal{A}(\mathcal{G}_{test},\mathcal{G}_{standard})==\text{True} \Longleftrightarrow \mathcal{P}_{test}\sim \mathcal{P}_{standard}$.
%\end{theorem}

Before establishing the proof, we first introduce the coloring refinement process of WL test for MILP/LP problem since it is the first step 1 in algorithm $\mathcal{A}$. 
For iteration $l$ of the algorithm we will be assigning to each node a tuple $H_{i}^L$ containing the node’s old compressed label and a multiset of the node’s neighbors' compressed labels. A multiset is a set (a collection of elements where order is not important) where elements may appear multiple times.

At each iteration $l$, we will additionally be assigning to each node a new “compressed” label $C^L_i$ with the same $H_{i}^L$ will get the same compressed label.

Repeat the above process for up to (m+n) (the number of nodes) iterations or until the partition of nodes by compressed label does not change from one iteration to the next, we will get a converged multiset.

In addition, we introduce preliminary tools for an algorithm-independent definition. 

In fact, unfoldable and symmetric decomposable can be defined without relying on WL-test algorithm. We introduced equivalent definitions based on stable partition index sets.

\begin{definition}[Stable Partition Index Sets]
\label{definition:stable_partition}
For a modeling instance $\mathcal{P}$ in the form of (\ref{equation:eq1}) with $n$ decision variables and $n$ constraints, define index set for optimization variables by $\mathcal{I} = \{I_1,I_2, \cdots, I_s\}$ and index set for constraints by $\mathcal{J} = \{J_1,J_2, \cdots, J_t\}$, where
\begin{itemize}
    \item $\bigcup_{l=1}^s I_l=\{1,2,\cdots, m\}$, $\bigcup_{k=1}^t J_k=\{1,2,\cdots,n\}$;
    \item $I_{l_i}\cap I_{l_j}=\emptyset$, $J_{k_p}\cap J_{k_q}=\emptyset,$ $\forall i,j \in [1,\cdots, |I_l|],i\neq j,\text{ and }p,q \in [1,\cdots, |J_k|],p\neq q.$
\end{itemize}
We say $(\mathcal{I},\mathcal{J})$ is a pair of stable partition index sets if the following condition holds:
\begin{enumerate}
    \item $(c_i,\tau_i) = (c_{i'},\tau_{i'}), \forall i,i' \in I_p$ for some $p\in{1,2,\cdots,s}$;
    \item $(b_j,\circ_j) = (b_{j'},\circ_{j'}), \forall j,j' \in J_q$ for some $q\in{1,2,\cdots,t}$;
    \item $\forall p \in {1,2,\cdots,s}, q\in{1,2,\cdots,t},$ and $i,i' \in I_p$, we have $\sum_{j\in J_q}a_{ij} = \sum_{j\in J_q}a_{i'j}$;
    \item $\forall p \in {1,2,\cdots,s},q\in{1,2,\cdots,t},$ and $j,j' \in J_q$, we have $\sum_{i\in I_p}a_{ij} = \sum_{i\in I_p}a_{ij'}$;
\end{enumerate}
\end{definition}

\begin{lemma}
\label{lemma:wl_stable}
If there are no collision of hash functions and their weighted averages, then WL test algorithm \ref{alg:WL_MILP} will finally terminated at some stable partition in $\mathcal{O}(m+n)$ iterations.
\end{lemma}
Lemma \ref{lemma:wl_stable} is proved in \cite{chen2022representing}.

\begin{definition}[Unfoldable, by trivial partition]
\label{definition:wl_det_partition}
$\mathcal{P}$ is unfoldable if $\exists$ stable partition index sets $\mathcal{I}$ and $\mathcal{J}$ such that $\mathcal{I}$ or $\mathcal{J}$ are trivial partitions, i.e. $s = m$ and $t = n$.
\end{definition}

\begin{definition}[Decomposable Symmetric, by grouped partition]
$\mathcal{P}$ is decomposable symmetric if the following condition holds: 

$\exists$ stable partition index set $\mathcal{I}$ and $\mathcal{J}$ such that:
\begin{enumerate}
    \item There are only two types of index set in $\mathcal{I}$ and $\mathcal{J}$. Type 1 set only contains a single index. Type 2 contains several indexes, denote type 2 sets by $I_1, \cdots, I_{s'}$; $J_1, \cdots, J_{t'}$. (By WL-test coloring, nodes with index in $I_i$ or $J_j$ share the same color.)
    \item Type 2 sets $I_1, \cdots, I_{s'}$ and $J_1, \cdots, J_{t'}$ are equal-sized with $|I_p| = |J_q| = k >1, \forall p \in \{1,2,\cdots, s'\}$ and $q \in \{1,2,\cdots, t'\}$. %for some integer constant $L=\frac{n}{s} = \frac{m}{t}$; 
    
    \item There exist k disjoint groups $S^1,\cdots, S^k$ such that $|S^i\cap I_p| = |S^i\cap I_p| = 1$; and $\forall a \in S^i, b \in S^j$ with $i\neq j$, $a$ disconnected with $b$. 
\end{enumerate}
\end{definition}
By Lemma \ref{lemma:wl_stable}, we can show two sets of definitions are equivalent.

\subsection{Proof of Theorem \ref{theorem:sufficient_equal_detect}}
We construct the proof by two lemmas to illustrate sufficient conditions that the result of WL test coloring can reliably infer graph isomorphism.
\begin{lemma}
\label{lem:sufficient_one}
Suppose $\mathcal{P}_{standard}$ is unfoldable, then $\mathcal{G}_{standard}$ and $\mathcal{G}_{test}$ shares the same coloring $\Longleftrightarrow$ $\mathcal{G}_{standard}\sim \mathcal{G}_{test}$.
\end{lemma}
Suppose $\mathcal{P}_{standard}$ is unfoldable, want to show $\mathcal{A}(\mathcal{G}_{test},\mathcal{G}_{standard})==\text{Equivalent} \Longleftrightarrow \mathcal{P}_{test}\sim \mathcal{P}_{standard}$.

If $\mathcal{P}_{test}\sim \mathcal{P}_{standard}$, it is trivial that $\mathcal{A}(\mathcal{G}_{test},\mathcal{G}_{standard})==\text{Equivalent}$.

Now, consider when $\mathcal{A}(\mathcal{G}_{test},\mathcal{G}_{standard})==\text{Equivalent}$ and $\mathcal{P}_{standard}$ unfoldable, we have $len(\mathbb{A}_1) = len(\mathcal{C}_1) \And len(\mathbb{A}_2) = len(\mathcal{C}_2)$. 

By the detection algorithm, every color in the multisets output by WL test must be distinct, and multisets for $\mathcal{P}_{standard}$ are the same as multisets for $\mathcal{P}_{standard}$. 

One stable partition of $\mathcal{G}_{standard}$ and is $\{I_1,\cdots,I_n\}, \{J_1,\cdots,J_m\}$, where $I_k, J_l$ are single-element sets. WLOG, assume $I_k = i_k, J_l = j_l$. 

Similarly, denote the stable partition of $\mathcal{G}_{test}$ by $\{I'_1,\cdots,I'_n\}, \{J'_1,\cdots,J'_m\}$, with $I'_k = [i'_k], J'_l = [j'_l]$.

Now, define a bijection mapping that shuffles $[i_1, \cdots, i_m]$ and $[j_1,\cdots, j_n]$ to get $[i'_1, \cdots, i'_m]$ and $[j'_1,\cdots, j'_n]$, denote such mapping by $\mathbf{P}$. (Since each element in $[i_1, \cdots, i_m], [j_1,\cdots, j_n], [i'_1, \cdots, i'_m],$ or $[j'_1,\cdots, j'_n]$ is distinct, we can uniquely find such bijection). 

Notice that such bijection may only map the index of $v_i^{standard}$ to the index of $v_j^{test}$ and map the index of $w_l^{standard}$ to the index of $w_p^{test}$, we can separately define a bijection for decision variable index as $\mathbf{P}_1$ and a bijection for constraint index as $\mathbf{P}_2$.

Therefore, exists bijection $\mathbf{P}_1$ and $\mathbf{P}_2$ such that $\mathcal{P}_{test}$ can be written in the following form:

\begin{align*}
\min_x& \hat{c}^Tx, \\
\text{s.t. }& \hat{A} x \hat{\circ} \hat{b}
\end{align*}

where $\hat b = P_2b_{standard}, \hat C = P_1C_{standard}, \hat A = P_2A_{standard}P_1 ,\,\hat\circ=P_2\circ_{standard}$. 
This implies $\mathcal{P}_{test} \sim \mathcal{P}_{standard}$.

\begin{lemma}
\label{lem:sufficient_two}
Suppose $\mathcal{P}_{standard}$,$\mathcal{P}_{test}$ are decomposible symmetric, then $\mathcal{G}_{standard}$ and $\mathcal{G}_{test}$ shares the same coloring $\Longleftrightarrow$ $\mathcal{G}_{standard}\sim \mathcal{G}_{test}$.
\end{lemma}
% share the same coloring set after WL test.

%%%%%%%%%% 说明multiset index相同时颜色相同

When $\mathcal{P}_{standard}$ is decomposible symmetric, and algorithm $\mathcal{A}$ output "Equivalent", the partition sets of $\mathcal{G}_{standard}$ and $\mathcal{G}_{test}$ can be denoted as 
$$
\begin{aligned}
\mathcal{I}_{standard} &= [I_1,\cdots,I_k,I_{k+1},\cdots, I_s];\\
\mathcal{J}_{standard} &= [J_1,\cdots,J_l,J_{l+1},\cdots,J_t];\\
\mathcal{I}_{test} &= [\hat I_1,\cdots,\hat I_k,\hat I_{k+1},\cdots, \hat I_s];\\
\mathcal{J}_{test} &= [\hat J_1,\cdots,\hat J_l,\hat J_{l+1},\cdots,\hat J_t],
\end{aligned}
$$
where set $[I_1,\cdots,I_k],[\hat I_1,\cdots,\hat I_k], [J_1,\cdots,J_l],[\hat J_1,\cdots,\hat J_l],$ only contains one index, and set\\ $[I_{k+1},\cdots,I_s],[\hat I_{k+1},\cdots,\hat I_s], [J_{k+1},\cdots,J_t],[\hat J_{k+1},\cdots,\hat J_t]$ consist at least 2 indexes; $I_{i}$, $\hat{I}_i$ shares the same color $\forall i$; and $J_{j}$, $\hat{J}_j$ shares the same color $\forall j$.
\begin{comment}
By the definition of decomposable symmetric instances, for any two sets \\$K, S \in [I_{k+1},\cdots, I_s, J_{k+1},\cdots, J_t]\text{ or } [\hat I_{k+1},\cdots,\hat I_s, \hat J_{k+1},\cdots,\hat J_t]$,  $K$ and $S$ are either disconnected or exists a bijection connection between nodes from $K$ to $S$.
\end{comment}

Now, define a bijection mapping that maps 
$$[I_1,\cdots,I_k,I_{k+1},\cdots, I_s, J_1,\cdots,J_l,J_{l+1},\cdots,J_t]$$ to $$[\hat I_1,\cdots,\hat I_k,\hat I_{k+1},\cdots, \hat I_s, \hat J_1,\cdots,\hat J_l,\hat J_{l+1},\cdots,\hat J_t],$$
by the following rules:
\begin{enumerate}
\item For the unique index $i\in I_p$, where $p \in \{1,\cdots,k\}$, map $i$ to the unique index $i'\in \hat{I}_p$.
\item For the unique index $j\in J_q$, where $q \in \{1,\cdots,l\}$, map $j$ to to the unique index  $j'\in \hat{J}_q$.
\item For the remaining nodes, we consider a cluster-wise mapping, i.e, finding some equivalent clusters, mapping a cluster to another, and providing a unique mapping rule within chosen cluster. 

Let $V'$ and $\hat{V}'$ be sets of all variable nodes except those with unique color in $\mathcal{G}_{standard}$ and $\mathcal{G}_{test}$; $W'$ and $
\hat{W}'$ be sets of all constraint nodes except those with unique color in $\mathcal{G}_{standard}$ and $\mathcal{G}_{test}$.

Find clusters $S^1,\cdots, S^r$ such that each $S^i$ is disconnected, disjoint, consists of nodes with the same combination of unique colors as other $S^i$, and $\bigcup_{i=1}^{r}S^i = V'\cup W'$. Similarly, for symmetric decomposable $\mathcal{G}_test$ with the same coloring distribution, we can find clusters $\hat{S}^1,\cdots, \hat{S}^r$ such that $\hat{S}^i$ has the same coloring distribution with $S^i$, and each $S^i$ is disconnected, disjoint, consists of nodes with the same combination of unique colors as other $S^i$, and $\bigcup_{i=1}^{r}\hat{S}^i = 
\hat{V}'\cup \hat{W}'$. 

The existence of $S^1,\cdots, S^r$ and $\hat{S}^1,\cdots, \hat{S}^r$ are guaranteed by the symmetric decomposable property of $\mathcal{G}_{standard}$ and $\mathcal{G}_{test}$.

Now, we can define a bijection that maps $S^i$ to a corresponding cluster $\hat{S}_{i}$. Note that nodes in one cluster have distinct colors. The bijection mapping maps nodes from cluster $S^i$ to $\hat{S}_{i}$ according to color-matching, i.e. a node maps to another one when they are in the same color.
\end{enumerate}

Now, consider the adjacency matrix of the representing bipartite graph 
$$
\mathbf{A}_{adj} =\begin{bmatrix}0&\mathbf{A}^T\\\mathbf{A}&0\end{bmatrix}.
$$
Since node groups $S^1, \cdots, S^k$ are disconnected, we can rearrange matrix $A$ by some column permutation $\mathbf{P}^b_1$ and row permutation $\mathbf{P}^b_2$ such that 
$$
\mathbf{P}^b_1\mathbf{A}\mathbf{P}^b_2 = 
\begin{bmatrix}
\mathbf{A}_{1}&0 & \cdots & 0&\mathbf{a_1} \\
0&\mathbf{A}_2&\cdots & 0&\mathbf{a_2} \\
\vdots&\vdots &\ddots&\vdots & \vdots \\
0&0 & \cdots &\mathbf{A}_r&\mathbf{a_r} \\
\mathbf{b_1}^T&\mathbf{b_2}^T & \cdots&\mathbf{b_r}^T &\mathbf{A}_{r+1}\\ 
\end{bmatrix}\\
$$
where $A_1,\cdots,A_r$ are coefficient matrix for r clusters $S^1,\cdots,S^r$ with associated decision variables and constraints, and $A_{r+1}$ is a $k \times l$ matrix.

The above composition of bijection mapping operations is equivalent to applying permutation operations on $A_{standard}, b_{standard}, c_{standard}, \circ_{standard}$ by the following steps:
\begin{enumerate}
\item point-wise mapping for variables: permute $c$ and $A$ by permutation matrix $\mathbf{P}^0_1$ to map unique index $i\in I_p$ to $i'\in \hat{I_p}$, which produce $\hat{c} = \mathbf{P}^0_1c_{standard}$ and $\hat{A} = A_{standard}\mathbf{P}^0_1$
\item point-wise mapping for constraints: permute $b_{standard},\circ_{standard}$ and $\hat{A}$ by permutation matrix $\mathbf{P}^0_2$ to map unique index $j\in I_q$ to $j'\in \hat{J_q}$, which produce $\hat{b} = \mathbf{P}^0_2b_{standard}$, $\hat{\circ} = \mathbf{P}^0_2\circ_{standard}$ and $\hat{A} = \mathbf{P}^0_2\hat{A} = \mathbf{P}^0_2A_{standard}\mathbf{P}^0_1$
\item clustering mapping: permute $\hat{c}$ and $\hat{A}$ by permutation matrix $\mathbf{P}^c_1$ and permute $\hat{b},\hat{\circ}$ and $\hat{A}$ by permutation matrix $\mathbf{P}^c_2$ to produce
$$
\begin{aligned}
\hat{A} &= \mathbf{P}^c_1\hat{A}\mathbf{P}^c_2 = 
\begin{bmatrix}
\mathbf{A}_{1}&0 & \cdots & 0&\mathbf{a_1} \\
0&\mathbf{A}_2&\cdots & 0&\mathbf{a_2} \\
\vdots&\vdots &\ddots&\vdots & \vdots \\
0&0 & \cdots &\mathbf{A}_r&\mathbf{a_r} \\
\mathbf{b_1}^T&\mathbf{b_2}^T & \cdots&\mathbf{b_r}^T &\mathbf{A}_{r+1}\\ 
\end{bmatrix}\\ 
& = \mathbf{P}^c_1\mathbf{P}^0_1A_{standard}\mathbf{P}^0_2\mathbf{P}^c_2,\\
\end{aligned}
$$
$\hat{b} = \mathbf{P}^c_2\mathbf{P}^0_2b_{standard}$, $\hat{\circ} = \mathbf{P}^c_2\mathbf{P}^0_2\circ_{standard}$, and $\hat{c} = \mathbf{P}^c_1\mathbf{P}^0_1c_{standard}$
\item in-cluster mapping: iteratively permute $\hat{c}$ and $\hat{A}$ by permutation matrices $\mathbf{P}^1_1, \cdots, \mathbf{P}^r_1$ and permute $\hat{b},\hat{\circ}$ and $\hat{A}$ by permutation matrices $\mathbf{P}^1_2, \cdots, \mathbf{P}^r_2$ to produce $\hat{A} = \mathbf{P}^r_1\cdots \mathbf{P}^1_1\mathbf{P}^c_1\mathbf{P}^0_1A_{standard}\mathbf{P}^0_2\mathbf{P}^c_2\mathbf{P}^1_2\cdots \mathbf{P}^r_2$, $\hat{b} = \mathbf{P}^k_2\cdots\mathbf{P}^1_2\mathbf{P}^b_2\mathbf{P}^0_2b_{standard}$, $\hat{\circ} = \mathbf{P}^r_2\cdots\mathbf{P}^1_2\mathbf{P}^b_2\mathbf{P}^0_2\circ_{standard}$, and $\hat{c} = \mathbf{P}^r_1\cdots \mathbf{P}^1_1\mathbf{P}^c_1\mathbf{P}^0_1c_{standard}$

\end{enumerate}

Now, define $\mathbf{P}_1 = \mathbf{P}^k_1\cdots \mathbf{P}^1_1\mathbf{P}^c_1\mathbf{P}^0_1$ and $\mathbf{P}_2 = \mathbf{P}^k_2\cdots \mathbf{P}^1_2\mathbf{P}^c_2\mathbf{P}^0_2$, we can write $\mathcal{P}_{test}$ in the following form:

\begin{align*}
\min_x& \hat{c}^Tx, \\
\text{s.t. }& \hat{A} x \hat{\circ} \hat{b}
\end{align*}

where $\hat b = P_2b_{standard}, \hat C = P_1C_{standard}, \hat A = P_2A_{standard}P_1 ,\,\hat\circ=P_2\circ_{standard}$. 
This implies $\mathcal{P}_{test} \sim \mathcal{P}_{standard}$.

%Algorithm \ref{alg:equal_detection} will permute the adjacency matrix of $\mathcal{G}_{standard}$ to be blockwise matrix. Each block represents a sub-graph of $\mathcal{G}_{standard}$. Since each subgraphs are WL-determinable, Algorithm \ref{alg:equal_detection} suffices to test isomorphism by checking whether $\mathcal{G}_{standard}$ can be decomposed to isomorphic subgraphs as well. 

% preliminaries, 定义wltest是什么。
% 可以放一些lemma
% 证明symmetric decomposable graphs 通过wltest之后 等价于给每个子图单独给

% 证明我们的算法能够判断是不是wl determinable & decomposable symmetric； 
% 

%\subsection{Other forms of definitions for WL-determinable and Symmetric Decomposable Instances}
%In the last section, we provide an algorithm-dependent definition. 

\subsection{Complexity Analysis}
\label{subsection: complexity}
For the two main types of problem realizations in our benchmark, Algorithm \ref{alg:WL_MILP} converges in $\mathcal{O}(m+n)$ iteration. In addition, for problems with $m$ variables and $n$ constraints, the time complexity to distinguish tested problem realizations from the standard realization is at most $\mathcal{O}(k(m+n)^2)$, which is is significantly lower than classical algorithms employed by popular solvers, such as simplex method for LP and branch and bound algorithm for MILP. Specifically, 
\begin{enumerate}
\item \textbf{For unfoldable problem instances}, algorithm \ref{alg:WL_MILP} converges in at most $\mathcal{O}(m+n)$ iterations according to lemma \ref{lemma:wl_stable}.
\item \textbf{For decomposable symmetric problem instances},
% By \cite{chen2022representing} Theorem A.2,for any bi-graph $\mathcal{G}$, after $\mathcal{O}(\verb|V|+\verb|W|)$ 
algorithm \ref{alg:WL_MILP} converges in at most $\mathcal{O}(m+n)$ iterations, and we shall further conduct symmetric decomposable detection using algorithm \ref{alg:sym_decomp}, which takes time complexity $\mathcal{O}(kmn)$ in the worst case, where $k$ is the number of clusters in the symmetric decomposable graph. The total time complexity could be $\mathcal{O}(kmn)$.
%$\mathcal{O}(\verb|V| \times \verb|W|+\verb|V|\times\log\verb|V|)$.
\end{enumerate}

\subsection{Randomly sampling suffices to obtain symmetric decomposable}
\label{appendix:sample}
To make WL test work, it is desirable to sample a symmetric decomposable instance. In Theorem \ref{theorem:thm_sample_continuous} and \ref{theorem:thm_sample_discrete}, we proved that for a large range of modeling problems with reasonable assumptions, we can sample a symmetric decomposable instance from its \textbf{parameter support} with probability 1. 

\begin{definition}[Modeling Parameter Support]
\label{appendix:para_set}
For a class of model formulation $\mathcal{M}$ with $n$ decision variables and $m$ constraints, the \textbf{parameter set} $\Theta(\mathcal{M})$ is a collection of all possible values for problem data $(\mathbf{A},\mathbf{c},\mathbf{b},\circ)$. The parameter set associated with decision variable $x_i$ is $\Theta(\mathcal{M},i) = \left\{[\mathbf{A}_{:,i}^T,c_i]\right\}$.
\end{definition}
An example of a formulation parameter support is attached in Appendix \cref{appendix: examples}. 
\begin{comment}
Given a model's parameter support, we say model $\mathcal{M}$ is a flexible model if, for any variables $x_i$ in $\mathcal{M}$, at least one of its associated parameters —whether the objective coefficient or any of the constraint coefficients —can be arbitrarily chosen from a sufficiently large parameter support. A formal definition of flexible model is as follows:
\begin{definition}[Flexible Model]
\label{definition:flexible_model}
We say a model $\mathcal{M}$ is \textbf{flexible} if the following condition holds:

$\forall$ variables $x_i, i= 1,\cdots n, \exists$ element $p \in [\mathbf{A}_{:,i}^T,c_i]$ s.t. $p$ can be arbitrarily chosen from some uncountable set $S(p)\subset \mathbb{R}$. In other words, for given model $\mathcal{M}$, for any variables $x_i$ in $\mathcal{M}$, at least one of its associated parameters —whether the objective coefficient or any of the constraint coefficients —can be arbitrarily chosen from a sufficiently large space.
\end{definition}
\end{comment}

\begin{theorem}[Efficient Sampling - continuous case]
\label{theorem:thm_sample_continuous}
Suppose a model $\mathcal{M}$ satisfies the following conditions: 

For each $\vec{\theta}_i \in \mathbb{R}^d, i=1, \cdots, n$, there exists a coordinate $k_i$ such that $\vec{e}_{k_i}^{\top} \vec{\theta}_i$ follows a continuous distribution $\mu_i$ independently across $i$.

then a random draw $\vec{\theta} \sim \Theta$ yields a \textbf{symmetric decomposable} instance $\mathcal{M}(\vec{\theta})$ almost surely.
\end{theorem}

\begin{theorem}[Efficient Sampling - discrete case]
\label{theorem:thm_sample_discrete}

Suppose a model $\mathcal{M}$ satisfies the following conditions:

$\forall i=1, \cdots, n, \forall \vec{\theta}_i \in \mathbf{R}^d, \exists k_i \in\{1,\cdots,d\}$ such that $\vec{e}_{k_i}^{\top} \vec{\theta}_i \sim \mu_i(\cdot)$ and independent of the distribution of $\vec{\theta}_j, \forall j\neq i$, where $\vec{e}_{k_i}$
is the $k_i$-th standard basis vector in $\mathbf{R}^d, \mu_i(\cdot)$ is some discrete uniform distribution with $u_i(\vec{e}_{k_i}^{\top} \vec{\theta}_i)_{\sim} Uniform\left\{x_1 \cdots  x_{l}\right\}$, i.e. at lease one coordinate of $\vec{\theta}_i$ can be randomly sampled with probability $\frac{1}{l}$, where $k_i$ is the index of coordinate in $\vec{\theta}_i$ that being sampled.

Then, as $l \rightarrow \infty$, randomly sample $\vec{\theta}$ from parameter support $\Theta$, we can get a symmetric decomposable instance for model $\mathcal{M}$ with probability 1.

\end{theorem}

We present the proof for Theorem \ref{theorem:thm_sample_continuous} and \ref{theorem:thm_sample_discrete} in Appendix \ref{appendix:proof_sample}.

\subsection{Proof of Theorem \ref{theorem:thm_sample_continuous} and Theorem \ref{theorem:thm_sample_discrete}}
\label{appendix:proof_sample}

\paragraph{Proof:}
\begin{lemma}
\label{lem:sample_continuous}
Suppose model $\mathcal{M}$ satisfies the following assuption: 

$\forall i=1, \cdots, n . \forall \vec{\theta}_i \in \mathcal{R}^d, \exists k_i \in \{1,\cdots,d\}$ such that $\vec{e}_{k_i}^{\top} \vec{\theta}_i \sim \mu_i\left(\vec{\theta}_i\right)$ and independent of the distribution of $\vec{\theta}_j$, where $\vec{e}_{k_i}$
is the $k_i$-th standard basis vertor in $\mathbf{R}^d$, $ \mu_i(\vec{\theta}_i)$ is some continuous distribution; i.e., at least one coordinate of $\vec{\theta}_i$ can be randomly sampled according to some continuous distribution. 

Then, we have $P(\vec{\theta}_i = \vec{\theta}_j) =0, \forall i\neq j$.
\end{lemma}

Proof of lemma \ref{lem:sample_continuous}:

Consider $i\neq j$, $0\leq P\left(\vec{\theta}_i=\vec{\theta}_j\right) \leq P\left(\vec{e}_{k_j} \vec{\theta}_j=\vec{e}_{k_j} \vec{\theta}_i\right)=0$ since $\mu_i$ is continuous distribution. 

By lemma \ref{lem:sample_continuous}, we have $P\left(\vec{\theta}_i \neq \vec{\theta}_j\right)=1$.
\begin{lemma}
\label{lem:sample_discrete}
Suppose model $\mathcal{M}$ satisfies the following condition:

$\forall i=1, \cdots, n, \forall \vec{\theta}_i \in \mathbf{R}^d, \exists k_i \in\{1,\cdots,d\}$ such that $\vec{e}_{k_i}^{\top} \vec{\theta}_i \sim \mu_i(\cdot)$ and independent of the distribution of $\vec{\theta}_j, \forall j\neq i$, where $\vec{e}_{k_i}$
is the $k_i$-th standard basis vector in $\mathbf{R}^d, \mu_i(\cdot)$ is some discrete uniform distribution with $u_i(\vec{e}_{k_i}^{\top} \vec{\theta}_i)_{\sim} Uniform\left\{x_1 \cdots  x_{l}\right\}$, i.e. at lease one coordinate of $\vec{\theta}_i$ can be randomly sampled with probability $\frac{1}{l}$, where $k_i$ is the index of coordinate in $\vec{\theta}_i$ that being sampled.

Then $P( \vec{\theta}_i=\vec{\theta}_i)\rightarrow 0 \text { as } l \rightarrow \infty$.
\end{lemma}
Proof of lemma \ref{lem:sample_discrete}:

$$
\begin{aligned}
 P\left( \vec{\theta}_i=\vec{\theta}_j\right) & =P\left(\vec{e}_{k_i} \vec{\theta}_i=\vec{e}_{k_i} \vec{\theta}_j\right) \\
&  =\sum_xP\left(\vec{e}_{k_i} \vec{\theta}_j = x \mid \vec{e}_{k_i}\vec{\theta}_i=x\right)P\left( \vec{e}_{k_i} \vec{\theta}_i=x\right) \\
& =\sum_xP\left(\vec{e}_{k_i} \vec{\theta}_j=x\right) \\
& =\sum_x\frac{1}{l^2} \\
& =  \frac{1}{l} .
\end{aligned}
$$

as $l \rightarrow \infty, \quad P\left(\vec{\theta}_i=\vec{\theta}_j\right)\rightarrow 0$.

\begin{lemma}
\label{lem:sd_property}
Suppose a modeling instance $\mathcal{P}$ has $P(\vec{\theta}_j = \vec{\theta}_{j^{\prime}}) = 0, \forall j \neq j^{\prime}$, then 

$$P(\mathcal{P} \text{ is symmetric decomposable}) = 1$$.
\end{lemma}

Proof of lemma \ref{lem:sd_property}:

Suppose $\exists$ index set $k \subset\{1,2, \cdots, d\}$ such that $\forall k \in k, \vec{e}_k \vec{\theta}_j \neq \vec{e}_k \vec{\theta}_j$, want to show the joint probability of the following event is 1 :
\begin{enumerate}
    \item Event A: $c_j \neq c_{j^{\prime}}$. [Objective coefficients are not the same.]
    \item Event B: $\sum_{i \in I} a_{i j} \neq \sum_{i \in I} a_{i j^{\prime}}$ [accumulated edge weights for variable nodes of $j,j^{\prime}$ are not the same].
    \item Event C: $\sum_{q \in J} a_{i^\prime q} \neq \sum_{q \in J}a_{iq}$ for some $J$ containing index $j$ or $j^{\prime}$ and
some $i \neq i^{\prime} \in I$. [accumulated edge weights for two constraint nodes are not the same];
\end{enumerate}
where $I$ and $J$ are sets in stable partitions $\mathcal{I},\mathcal{J}$.
It is equivalent to show $P(A\cup B\cup C) = 1$.
Now, consider two cases when $j,j^{\prime} \in \{1,\cdots,n\}$:
\begin{enumerate}
    \item Case 1: $\exists k \in K$ s.t. $\vec{e}_k^{\top}\vec{\theta}_j = c_j$, $\vec{e}_k^{\top}\vec{\theta}_{j^{\prime}} = c_{j^{\prime}}$, then $c_j\neq c_{j^{\prime}}$.
    \item Case 2: $\exists k \in K$ s.t. $\vec{e}_k^{\top}\vec{\theta}_j = a_{ij}$,$\vec{e}_k^{\top}\vec{\theta}_{j^{\prime}} = a_{ij^{\prime}}$ for some $i$, then $a_{ij}\neq a_{ij^{\prime}}$ for some $i$.
\end{enumerate}
Notice that $P(\Omega)=P(\text{Case 1} \cup \text{Case 2}) = 1)$.

It suffices to show $P(A \cup B \cup C \mid \text{case 1} \cup\text{case 2})=1$.
Now, $P(A \cup B \cup C \mid$ case 1$)=1$ since $P(A \mid$ case 1$)=1$.
It suffices to show $P(A \cup B \cup C \mid\text{case 2})=1$;
Is suffices to show $P(B \cup C \mid\text{case 2})=1$.

Now, suppose $\exists k \in K$ s.t. $\vec{e}_k\vec{\theta}_j=a_{i j} \neq \vec{e}_k \vec{\theta}_{j^{\prime}}=a_{i j^{\prime}}$.

Consider $I$ containing $i$. WLOG, suppose $\hat{I} \subset I$ is an index set that containing all $i's$ such that $a_{i j} \neq a_{i j^{\prime}}$, and $\sum_{i \in I / \hat{I}}\left(a_{i j}-a_{i j^{\prime}}\right)=c$ for come constant $c$,
then 

$$
\begin{aligned}
P\left(\sum_{i \in I} a_{i j} \neq \sum_{i \in I} a_{i j^{\prime}}\right)
&=P\left(I_{i \in \hat{I}} a_{i j} \neq \sum_{i \in I} a_{i j^{\prime}}+c\right)\\
& =1-P\left(I_{i \in \hat{I}} a_{i j} = \sum_{i \in I} a_{i j^{\prime}}+c\right) \\
& =1-0\\
&=1
\end{aligned}
$$

The third equality holds since $\sum_{i \in \hat{I}} a_{i j}$ and $\sum_{i \in \hat{I}} a_{i j^{\prime}}$ are independent and can be sampled from some continuous distribution. 
Therefore. $P(B \mid \text{case 2})=0$, we have $P(A \cup B \cup C) = P(A \cup B \cup C \mid \Omega) = P(A \cup B \cup C \mid (\text{case 1}\cup {case 2}))=1$.

$$
P(A \cup B \cup C)=P(A \cup B \cup C \mid \Omega)=P(A \cup B \cup C \mid \text { case } 1 \cup \text { case } 2)=1 \text {. }
$$

Now, by lemma \ref{lem:sample_continuous} and lemma \ref{lem:sd_property}, we can prove Theorem \ref{theorem:thm_sample_continuous}; by lemma \ref{lem:sample_discrete} and lemma \ref{lem:sd_property}, we can prove Theorem \ref{theorem:thm_sample_discrete}.

\section{Examples}
\label{appendix: examples}
\subsection{Examples for limitations of solver-based evaluation}
\begin{example}[The solver returns values, and the execution accuracy is 1 but the mathematical model is actually wrong]
\label{example:limitation1_example1}
Consider a car production and revenue maximization problem. A manufacturer produces two types of cars: sedans and SUVs. Let the decision variables of $x$ be the number of sedans to produce and $y$ be the number of SUVs to produce. The \underline{correct} formulation is:
\begin{align*}
\text{Maximize } & 30x + 50y  \\
\text{such that:} \,  & x + 2y \leq 100 \quad \text{(Production capacity in labor-hours)} \\
& x \geq 0, y \geq 0 \quad \text{(Non-negativity)}
\end{align*}
Now suppose an LLM generates an \underline{incorrect} model with an additional erroneous constraint:
\begin{align*}
\text{Maximize } & 30x + 50y  \\
\text{such that:} \, &x + 2y \leq 100 \quad \text{(Production capacity in labor-hours)}  \\
&x + y \leq 40 \quad \text{(ERRONEOUS constraint)} \\
&x \geq 0, y \geq 0 \quad \text{(Non-negativity)}
\end{align*}
If we test with a data configuration $\theta$ where production capacity = 80 and the market demand limit is 40, both models will yield the same optimal solution and optimal value: produce 40 SUVs for a revenue of \$2,000. However, if the data configuration changes to $\theta'$ with production capacity = 200, the correct model would recommend producing 100 SUVs for a revenue of \$5,000, while the incorrect model would still limit production to 40 units total due to the erroneous constraint.
\end{example}

\begin{example}[The solver returns constant value, and its useless for modeling equivalence detection]
\label{example:limitation1_example2}
Consider a facility location problem where the goal is to determine whether it is possible to open a subset of facilities to serve all customer demand within a fixed budget. The objective is a constant (e.g., 0), since only feasibility is of interest:

\begin{align*}
\text{Minimize } & 0 \\
\text{such that:} \quad & \sum_{j \in \mathcal{F}} x_j \cdot c_j \leq B \quad \text{(Budget constraint)} \\
& \sum_{j \in \mathcal{F}} a_{ij} x_j \geq 1 \quad \forall i \in \mathcal{D} \quad \text{(Coverage: each demand point must be served)} \\
& x_j \in \{0,1\} \quad \forall j \in \mathcal{F}
\end{align*}

Now, suppose the LLM generates an incorrect model with slightly relaxed constraints:

\begin{align*}
\text{Minimize } & 0 \\
\text{such that:} \quad & \sum_{j \in \mathcal{F}} x_j \cdot c_j \leq B \quad \text{(Same budget constraint)} \\
& \sum_{j \in \mathcal{F}} a_{ij} x_j \geq 0.5 \quad \forall i \in \mathcal{D} \quad \text{(ERRONEOUS weaker coverage)} \\
& x_j \in \{0,1\} \quad \forall j \in \mathcal{F}
\end{align*}

If both models happen to be feasible under a specific data configuration $\theta$ (e.g., a small number of facilities with low costs and high coverage), then the solver will return “feasible” for both. However, the second model allows partial coverage (due to the threshold of 0.5), which violates the intended semantics. Since the objective function is constant, execution accuracy based on solver output cannot detect this structural mistake.
\end{example}

\begin{example}[The mathematical model is incorrect but the execution accuracy is invalid for infeasible problems]
\label{example:limitation2_example1}
Consider the same car production problem, but with modified constraints:
\begin{align*}
\text{Maximize } & 30x + 50y  \\
\text{such that:} \,  & x + 2y \leq 10 \quad \text{(Limited production capacity in labor-hours)} \\
& x \geq 20, y \geq 0 \quad \text{(Minimum sedan production requirement)}
\end{align*}
This \underline{correct} model is genuinely infeasible because the minimum sedan production requirement ($x \geq 20$) cannot be satisfied with the limited production capacity ($x + 2y \leq 10$).
Now suppose an LLM generates an \underline{incorrect} model with an erroneous constraint:
\begin{align*}
\text{Maximize } & 30x + 50y  \\
\text{such that:} \, &x + 2y \leq 10 \quad \text{(Limited production capacity in labor-hours)}  \\
&x \geq 5, y \geq 7 \quad \text{(ERRONEOUS minimum requirements)} \\
&x, y \geq 0 \quad \text{(Non-negativity)}
\end{align*}
For the data configuration $\theta$ shown above, both models will be evaluated as infeasible by the solver. The execution accuracy metric cannot distinguish between the correct model that is genuinely infeasible under this configuration and the incorrect model that is infeasible due to contradictory constraints ($x \geq 5$ and $y \geq 7$ would require at least 19 labor-hours, exceeding the 10 available).
% If the resource constraint were relaxed in a new data configuration $\theta'$ (e.g., changing $x + 2y \leq 10$ to $x + 2y \leq 100$), the correct model would yield a feasible and optimal solution, while the incorrect model would also become feasible but produce a suboptimal solution due to the erroneous minimum requirements.
\end{example}

\subsection{Examples for model, model parameter set, and model instance}
\begin{example}[Model Parameter Set for Blending Problem]
\label{example:model_parameter_set}
For example, a blending problem can be formulated as:
$$
\begin{aligned}
\min_x &\sum_{i=1}^n c_ix_i\\
\text{s.t. } & \sum_{i=1}^n a_{ji}x_i \geq p_j, \forall j = 1, \cdots, m.\\
& x_i \leq u_i, \forall i = 1, \cdots, n.
\end{aligned}
$$
The corresponding parameter set $\Theta(\mathcal{M}_{blend})$ can be defined as 
$$
\begin{aligned}
\Theta(\mathcal{M}_{blend}) &= \Big\{
(\mathbf{A},\mathbf{c},\mathbf{b},\circ)\Big|
\mathbf{A} =[\hat{\mathbf{A}}^T,I_n]^T, \text{ where }\hat{\mathbf{A}}\in\mathbb{R}^{m\times n} \text{ and } I_{n} \text{ is an }n\times n \\
&\text{ identity matrix}; \mathbf{c} = [c_1,\cdots,c_n]^T \in\mathbb{R}^n; \mathbf{b} = [-p_1,\cdots, -p_J,-u_1, \cdots, -u_n]^n\in \mathbb{R}^{m+n};\\
&\circ = [\geq, \cdots,\geq,\leq,\cdots, \cdots, \leq]^T_{1\times(m+n)}
\Big\}.
\end{aligned}
$$
The parameter set associated with $x_i$ is $\Theta(\mathcal{M}_{blend},i) = \left\{[\mathbf{A}_{:,i}^T,c_i]\right\} = \mathbb{R}^{m+1}$.
\end{example}
\begin{figure}[t]
\label{figure:model_instance_generation}
\begin{center}
%\framebox[4.0in]{$\;$}
%\fbox{\rule[-.5cm]{0cm}{4cm} \rule[-.5cm]{4cm}{0cm}}
\includegraphics[width = \textwidth]{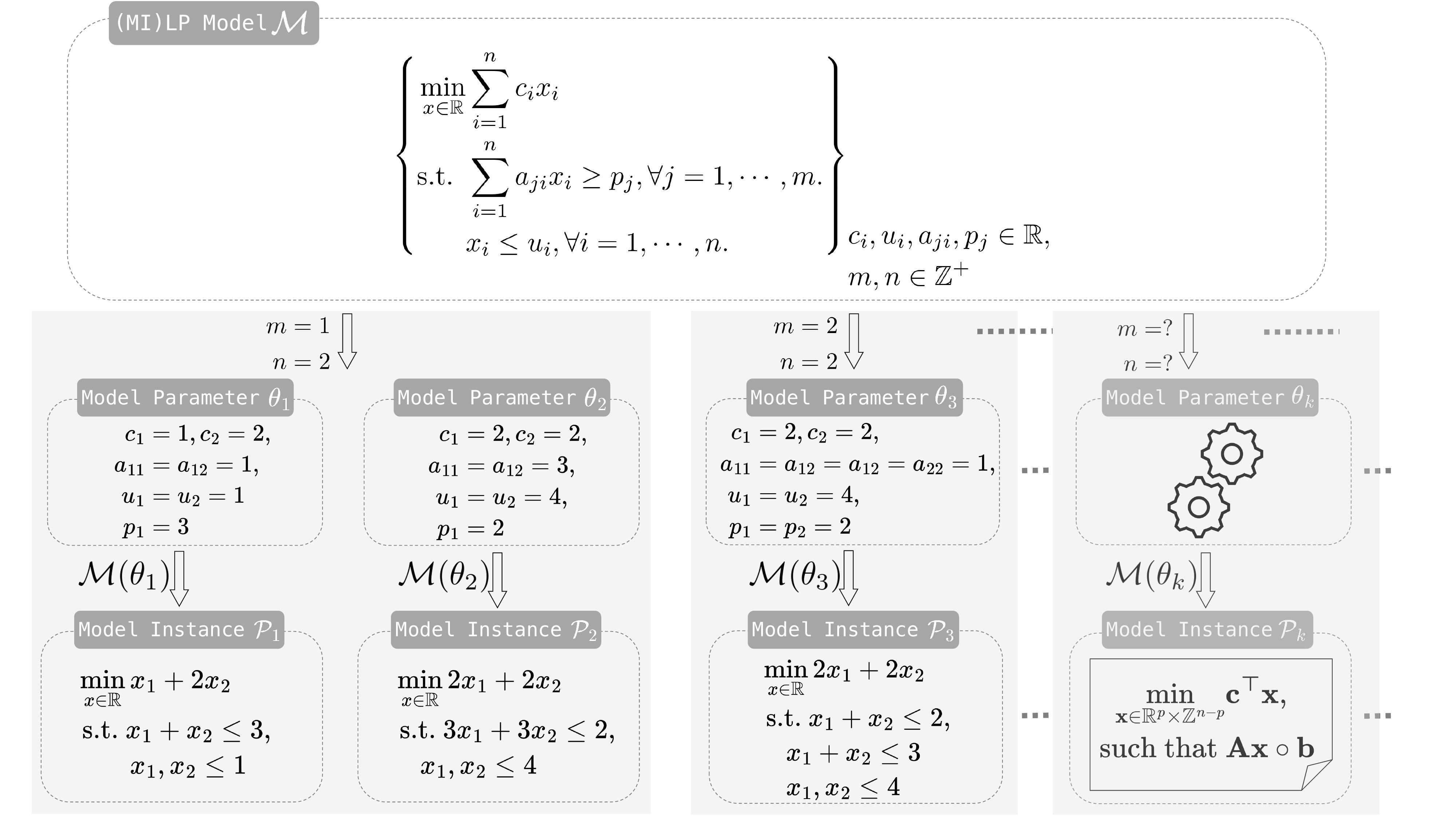}
\end{center}
\caption{Model Instances Generation}
\end{figure}

\subsection{Examples for Symmetry}
\label{example:symmetric_example}
\begin{example}[Undesirable Symmetry]
Discriminating problem instances involving symmetry in their decision variables or constraints can be tricky. Because some non-isomorphic bipartite graphs cannot be distinguished by WL-test due to their automorphic structure in the graph. For example, \cite{chen2022representing} illustrates one case in which two MILP graphs are non-isomorphic while WL-test outputs the same multiset.

\begin{figure}[H]
\begin{center}
%\framebox[4.0in]{$\;$}
%\fbox{\rule[-.5cm]{0cm}{4cm} \rule[-.5cm]{4cm}{0cm}}
\includegraphics[width=1\linewidth]{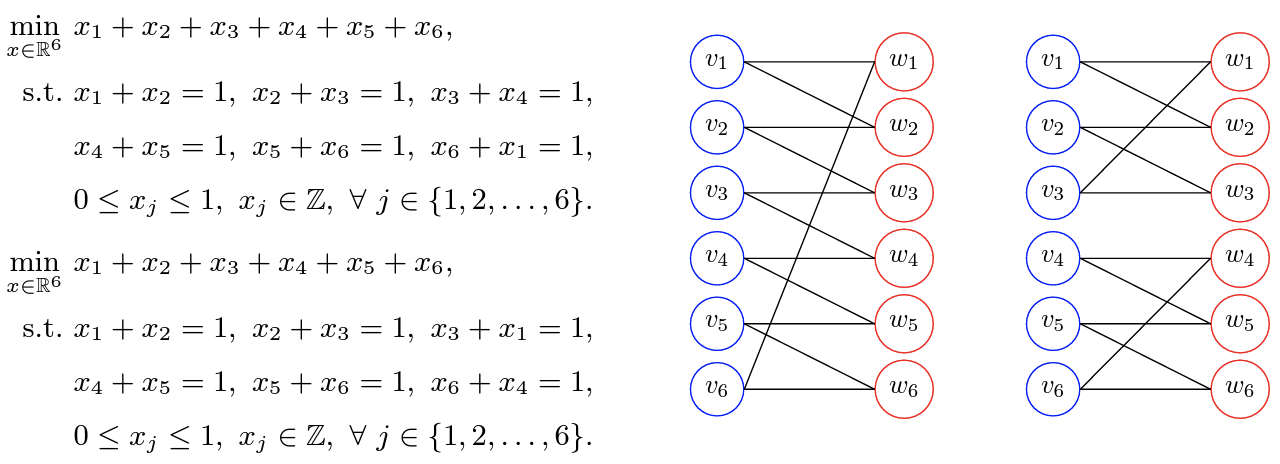}
\label{fig:symmetry_example}
\end{center}
\caption{Two non-isomorphic MILP graphs that cannot be distinguished by WL test}
\end{figure}

\end{example}

\begin{figure}[H]
\begin{center}
%\framebox[4.0in]{$\;$}
%\fbox{\rule[-.5cm]{0cm}{4cm} \rule[-.5cm]{4cm}{0cm}}
\includegraphics[width = 0.5\textwidth]{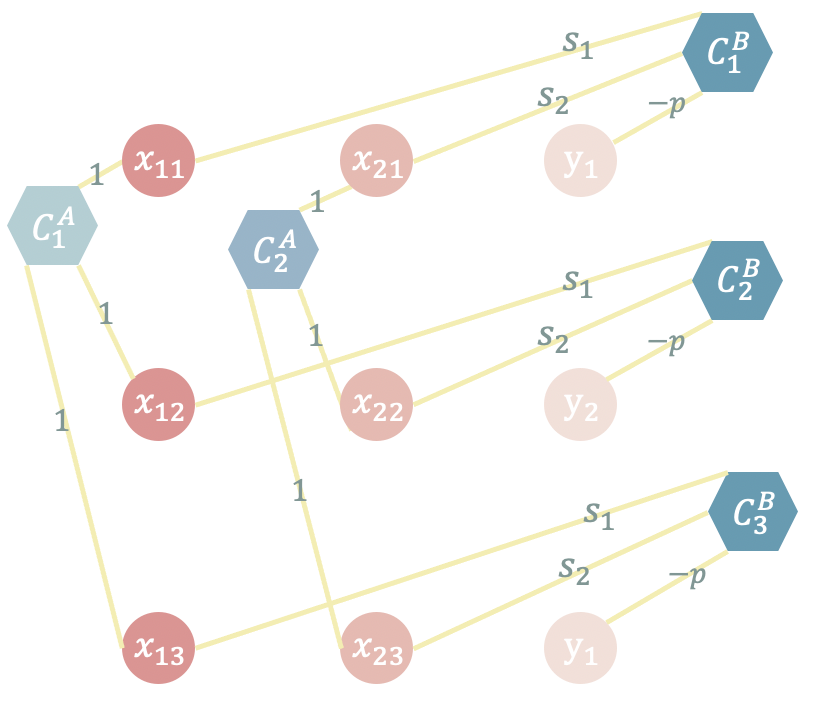}
\end{center}
\caption{Bipartite for a bin-packing problem. Different colors indicate that the nodes are colored using the WL test. This figure illustrates the representation of a symmetric decomposable graph. There are four groups of nodes with the same colors in each group, and two nodes with distinct colors. In addition, a node in any group, for example, the lightest red group, only connects with one node in other groups. }
\label{fig:decomp_symmetric}
\end{figure}

\begin{example}[Symmetric Decomposable Problem]
For decomposable symmetric problems, their corresponding bipartite graph can be divided into several symmetric sub-graphs, with each isomorphic and disconnected from others. For example, an instance on bin-packing with heterogeneous vehicles is formulated as
$$
\begin{aligned}
\min_{x\in\{0,1\}^q,y\in\{0,1\}^p} &\sum_{j=1}^p y_j\\
\text{s.t. } & \sum_is_ix_{ij} \leq by_j, \forall j = 1, \cdots, p.\\
& \sum_{j=1}^p x_{ij} = 1, \forall i = 1,\cdots,q
\end{aligned}
$$
For the bin-packing problem with $p=3$ and $q=2$, a corresponding bipartite is illustrated in figure \ref{fig:decomp_symmetric}, where the red node represents decision variables and the blue nodes represent constraints.
\end{example}

\begin{figure}[H]
\begin{center}
%\framebox[4.0in]{$\;$}
%\fbox{\rule[-.5cm]{0cm}{4cm} \rule[-.5cm]{4cm}{0cm}}
\includegraphics[width = 0.5\textwidth]{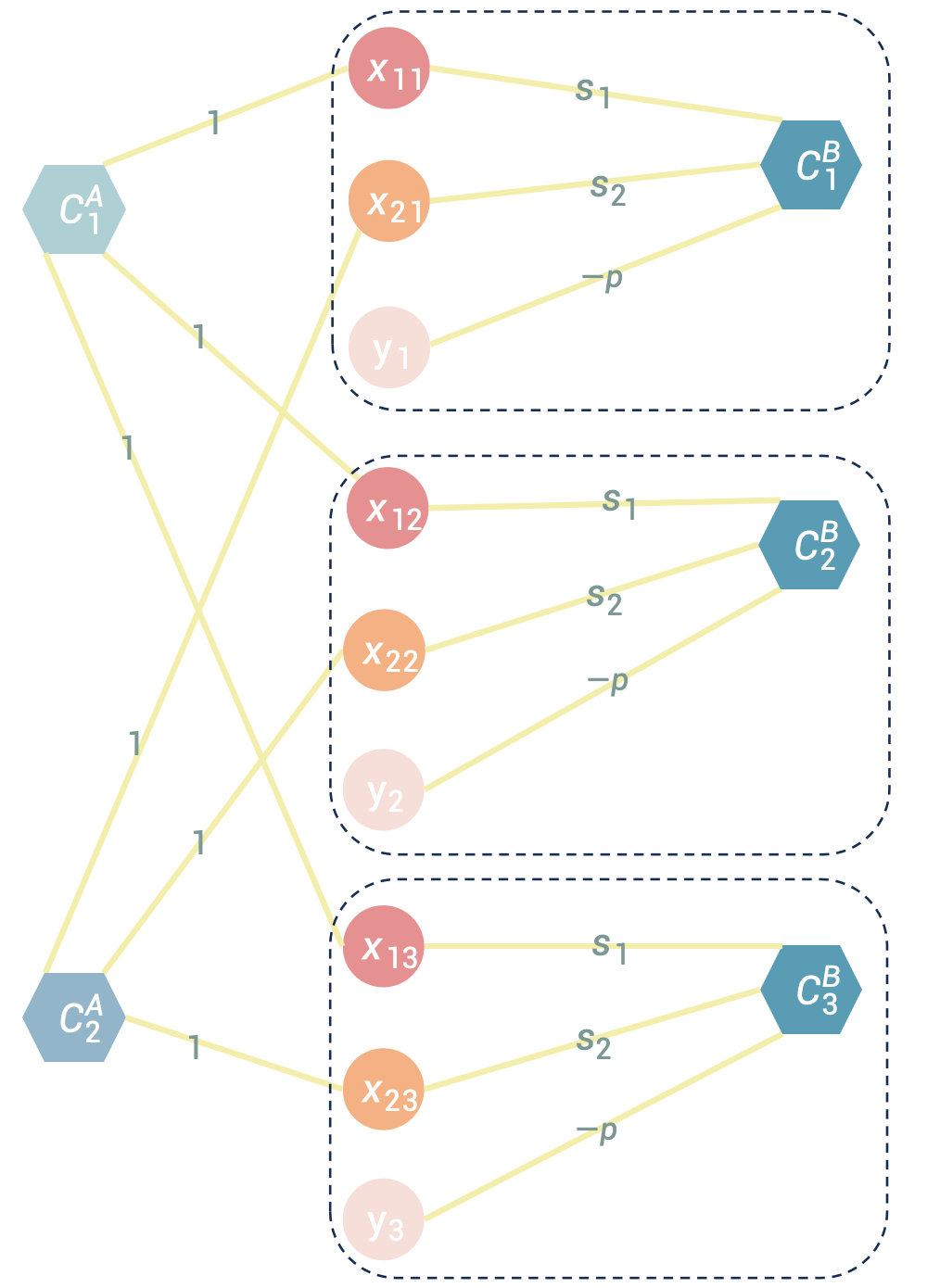}
\end{center}
\caption{Decompose a symmetric decomposable graph}
\label{fig:symmetric_decomposable}
\end{figure}

Such graphs are quite special since by excluding uniquely colored nodes and their connecting edges, the remaining symmetric nodes (nodes labeled in the same color via the WL test) can be combined to form several isomorphic, disconnected, and unfoldable graphs, as the dashed line highlights in Figure \ref{fig:symmetric_decomposable}.

\newpage
\begin{example}[Solver can be Inconsistant]
\label{example:solver_inconsistant}
\end{example}
\begin{figure}[ht]
\begin{center}
%\framebox[4.0in]{$\;$}
%\fbox{\rule[-.5cm]{0cm}{4cm} \rule[-.5cm]{4cm}{0cm}}
\includegraphics[width = \textwidth]{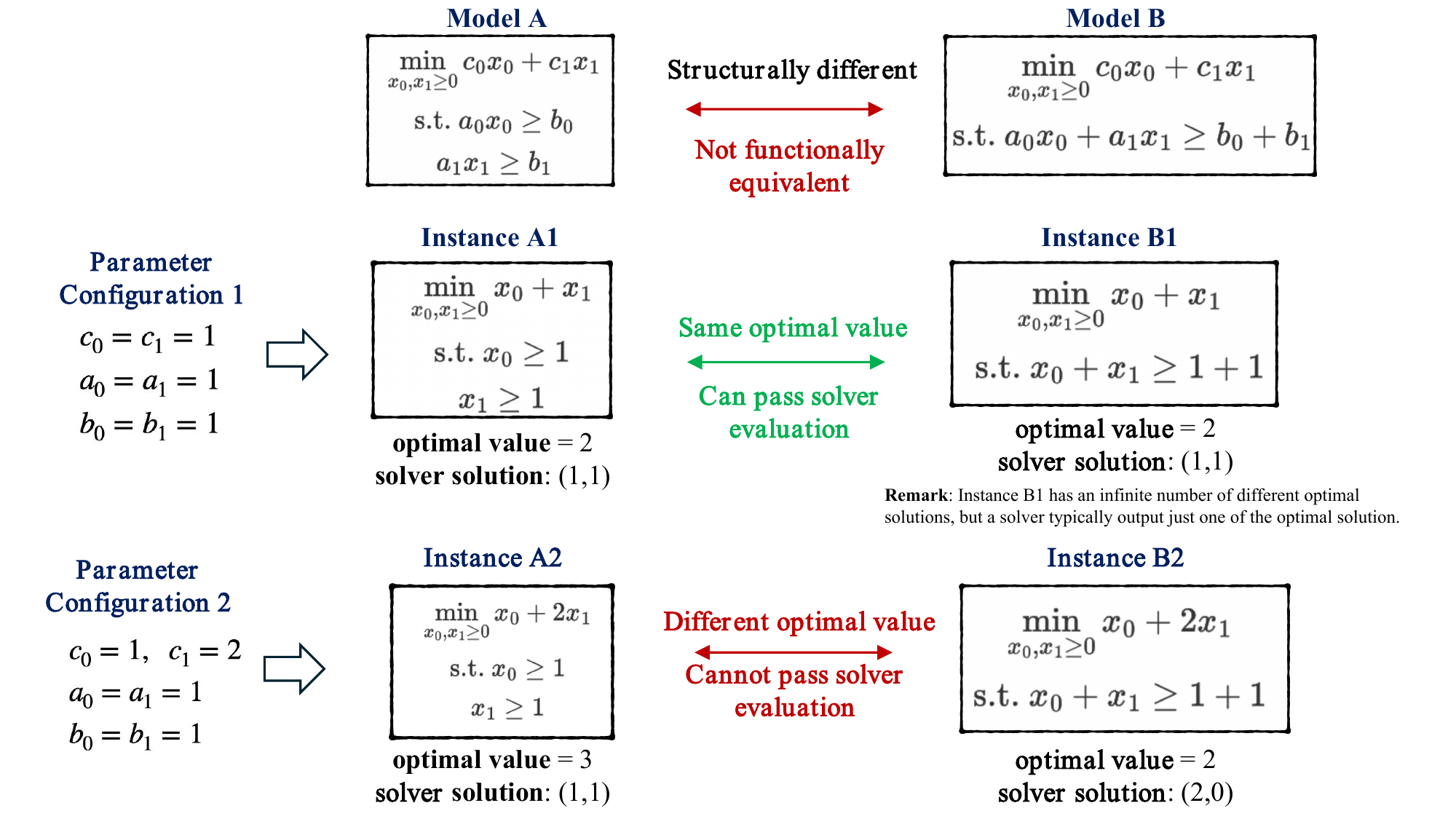}
\end{center}
\caption{Solver's evaluation on modeling equivalence can be inconsistent across different parameter configurations}
\label{fig:solver_inconsistant_example}
\end{figure}

\end{document}